\documentclass{article}

\PassOptionsToPackage{numbers, sort&compress}{natbib}

\usepackage[preprint]{neurips_2021}

\usepackage[utf8]{inputenc} %
\usepackage[T1]{fontenc}    %
\usepackage{booktabs}       %
\usepackage{amsfonts}       %
\usepackage{nicefrac}       %
\usepackage{microtype}      %

\usepackage{hyperref}

\usepackage{amsmath, amssymb,amsthm}
\usepackage{graphicx}
\usepackage{bbm}
\usepackage{subcaption}
\usepackage[font=small]{caption}

\usepackage{todonotes}
\usepackage{enumitem}
\usepackage{float}
\usepackage{algorithm}
\usepackage{booktabs}
\usepackage{colortbl}
\usepackage{xcolor}
\usepackage{multirow}

\usepackage{titletoc}
\usepackage{hyperref}
\usepackage[page,header]{appendix}
\usepackage[T1]{fontenc}    %
\usepackage{url}            %
\usepackage{booktabs}       %
\usepackage{amsfonts}       %
\usepackage{nicefrac}       %
\usepackage{microtype}      %
\usepackage{xcolor}
\usepackage{hyperref}
\usepackage{graphicx}
\usepackage{algorithmicx}
\usepackage{algorithm}%
\usepackage{algpseudocode}%
\usepackage{mathrsfs}
\usepackage{amssymb}
\usepackage{graphicx,wrapfig,lipsum}
\usepackage{caption}
\usepackage{float}
\usepackage{subcaption}
\usepackage{enumitem}
\usepackage{siunitx}
\usepackage{multirow}
\usepackage{tikz}
\usepackage{footnote}

\usepackage{lipsum}
\usepackage{frame}
\usepackage[framemethod=TikZ]{mdframed} 
\usepackage[multiple]{footmisc}
\usepackage{placeins}

\usepackage{url}

\definecolor{mydarkblue}{rgb}{0,0.08,0.45}
\hypersetup{ %
    pdftitle={},
    pdfauthor={},
    pdfsubject={},
    pdfkeywords={},
    pdfborder=0 0 0,
    pdfpagemode=UseNone,
    colorlinks=true,
    linkcolor=mydarkblue,
    citecolor=mydarkblue,
    filecolor=mydarkblue,
    urlcolor=mydarkblue,
    pdfview=FitH}

\usepackage{listings}
\usepackage{color}
\definecolor{codegreen}{rgb}{0,0.5,0}
\definecolor{codeblue}{rgb}{0,0,0.9}
\definecolor{codegray}{rgb}{0.5,0.5,0.5}
\definecolor{codepurple}{rgb}{0.58,0,0.82}
\definecolor{backcolour}{rgb}{0.95,0.95,0.92}
\definecolor{backcolour2}{rgb}{0.9,0.9,0.9}
\definecolor{codered}{rgb}{0.5,0,0}
\definecolor{textcodered}{rgb}{0.4,0,0}
\definecolor{palegray}{rgb}{0.98,0.98,0.99}

\lstdefinestyle{mystyle}{
    backgroundcolor=\color{backcolour},
    commentstyle=\color{codegreen},
    keywordstyle=\color{codeblue},
    numberstyle=\tiny\color{codegray},
    stringstyle=\color{codegreen},
    breakatwhitespace=false,
    breaklines=true,
    captionpos=b,
    keepspaces=true,
    numbersep=5pt,
    showspaces=false,
    showstringspaces=false,
    showtabs=false,
    tabsize=2,
    otherkeywords={with},
    basicstyle=\ttfamily\scriptsize
}

\definecolor{Gray}{gray}{0.90}
\definecolor{White}{RGB}{255,255,255}
\newcolumntype{g}{>{\columncolor{Gray}}c}
\definecolor{ffe1da}{RGB}{255,225,218}
\definecolor{F7E0D5}{RGB}{247,224,213}
\definecolor{40E0D0}{RGB}{175,238,238}
\definecolor{darkF7E0D5}{RGB}{209,154,128}
\colorlet{Light}{backcolour}

\title{Isaac Gym: High Performance GPU-Based Physics Simulation For Robot Learning}

\author{%
Viktor Makoviychuk, 
Lukasz Wawrzyniak, 
Yunrong Guo, 
Michelle Lu, 
Kier Storey, \And
Miles Macklin,
David Hoeller,
Nikita Rudin,
Arthur Allshire, 
Ankur Handa, 
Gavriel State\\ \\
NVIDIA \\
\texttt{\{vmakoviychuk, 
lwawrzyniak, kellyg, michellel, kstorey, mmacklin,} \\ 
\texttt{dhoeller, nrudin, aallshire, ahanda, gstate\}@nvidia.com}
}

\begin{document}

\maketitle

\begin{abstract}
  Isaac Gym offers a high performance learning platform to train policies for wide variety of robotics tasks directly on GPU. Both physics simulation and the neural network policy training reside on GPU and communicate by directly passing data from physics buffers to PyTorch tensors without ever going through any CPU bottlenecks. This leads to blazing fast training times for complex robotics tasks on a single GPU with 2-3 orders of magnitude improvements compared to conventional RL training that uses a CPU based simulator and GPU for neural networks. We host the results and videos at \url{https://sites.google.com/view/isaacgym-nvidia} and isaac gym can be downloaded at \url{https://developer.nvidia.com/isaac-gym}.
\end{abstract}

\clearpage
\tableofcontents
\clearpage

\section{Introduction}

\begin{figure*}[ht]
\includegraphics[width=\linewidth]{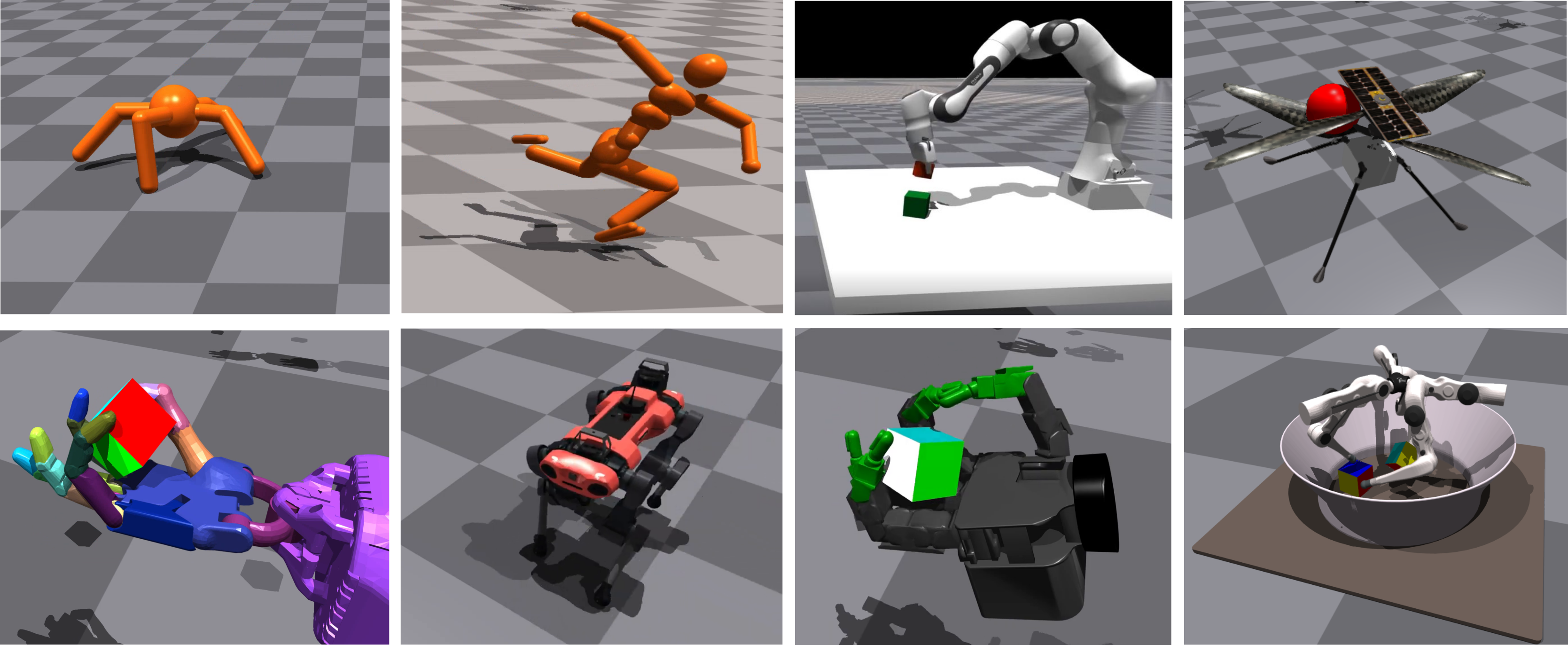}
\caption{Isaac Gym allows high performance training on a variety of robotics environments. We benchmark on 8 different environments that offer a wide range of complexity and show the strengths of the simulator in blazing fast policy training on a single GPU. \textit{Top}: Ant, Humanoid, Franka-cube-stack, Ingenuity. \textit{Bottom}: Shadow Hand, ANYmal, Allegro, TriFinger.}
\label{fig:hand-environments}
\end{figure*}

In recent years, reinforcement learning (RL) has become one of the most promising research areas in machine learning and has demonstrated great potential for solving sophisticated decision-making problems. Deep reinforcement learning (Deep RL) has achieved superhuman performance in very challenging tasks, ranging from classic strategy games such as Go and Chess \citep{Silver:etal:science2018}, to real-time computer games like StarCraft \citep{ontanon2013survey} and DOTA \citep{berner2019dota}. It has also shown impressive results in robotic settings, including legged locomotion \citep{RoboImitationPeng20} and dexterous manipulation \citep{openai-sh}.

Simulators play a key role in training robots improving both the safety and iteration speed in the learning process. Training a humanoid robot that walks up and down stairs in the real world can lead to damage to its machinery and the environment, including humans that are working on the robot. An alternative is to train inside simulators that offer an efficient and scalable platform via trial-and-error with no safety issues as observed in the real world. To date, most researchers have relied on a combination of CPUs and GPUs to run reinforcement learning system \cite{openai-sh}. Different parts of the computer tackle different steps of the physics simulation and rendering process. CPUs are used to simulate environment physics, calculate rewards, and run the environment, while GPUs are used to accelerate neural network models during training and inference as well as rendering if required.

However, switching back and forth between CPU cores optimized for sequential tasks and GPUs which offer large-scale parallelism is by nature inefficient, requiring data to be transferred between different parts of the system at multiple points during the training process. Therefore, scalability of deep reinforcement learning in robotics is faced with two critical bottlenecks: 1) enormous computational requirements and 2) limited simulation speed. These problems are especially challenging when learning long-horizon behaviours for robots with high degrees of freedom. 

Popular physics engines like MuJoCo\cite{MuJoCo:etal:2012}, PyBullet\cite{Pybullet:etal:2016}, DART\cite{Dart:etal:2018}, Drake\cite{Drake:etal:2019}, V-Rep\cite{Rohmer:etal:2013} \textit{etc.} need large CPU clusters to solve challenging RL tasks naturally face these bottlenecks. For instance, in \cite{akkaya2019solving}, almost 30,000 CPU cores (920 worker machines with 32 cores each) were used to train a robot to solve the Rubik’s Cube task using RL. In a similar task, \cite{openai-sh} used a cluster of 384 systems with 6144 CPU cores, plus 8 NVIDIA V100 GPUs, and required 30 hours of training for RL to converge. 
    
One way to speed-up simulation and training is to make use of hardware accelerators. GPUs have enjoyed enormous success in computer graphics are also naturally suited for highly parallel simulations. This approach was taken by \citep{Liang:etal:CoRL2018}, and showed very promising results running simulation on GPU, proving that it is possible to greatly reduce both training time as well as computational resources required to solve very challenging tasks using RL. However, some bottlenecks were still not addressed in the work -- simulation was on GPU but physics state was copied back to CPU. There, observations and rewards were calculated using optimized C++ code and later copied back to GPU where policy and value networks ran. Furthermore, only simplified physics-based scenarios were trained, rather than representative robotic environments, and no attempt was made to show sim2real.

To address these bottlenecks, we present \textbf{Isaac Gym} - an end-to-end high performance robotics simulation platform. It runs an end-to-end GPU accelerated training pipeline, which allows researchers to overcome the aforementioned limitations and achieves 2-3 orders of magnitude of training speed-up in continuous control tasks. Isaac Gym leverages NVIDIA PhysX  \cite{nvidia-physx} to provide a GPU-accelerated simulation back-end, allowing it to gather experience data required for robotics RL at rates only achievable using a high degree of parallelism. It provides a PyTorch tensor-based API to access the results of physics simulation natively on the GPU. Observation tensors can be used as inputs to a policy network and the resulting action tensors can be directly fed back into the physics system. We note that others \citep{brax2021github} have recently begun attempting an approach similar to Isaac Gym with respect to running end-to-end training on hardware accelerators.

With the end-to-end approach, roll-outs of observation, reward, and action buffers can stay on the GPU for the entire learning process, eliminating the need to read data back from the CPU. This set-up permits tens of thousands of simultaneous environments on a single GPU, allowing researchers to easily run experiments locally on their desktops that previously required an entire data center and to solve previously out of reach tasks using just a small GPU server.

\begin{figure}[h]
  \centering
  \includegraphics[width=.96\textwidth]{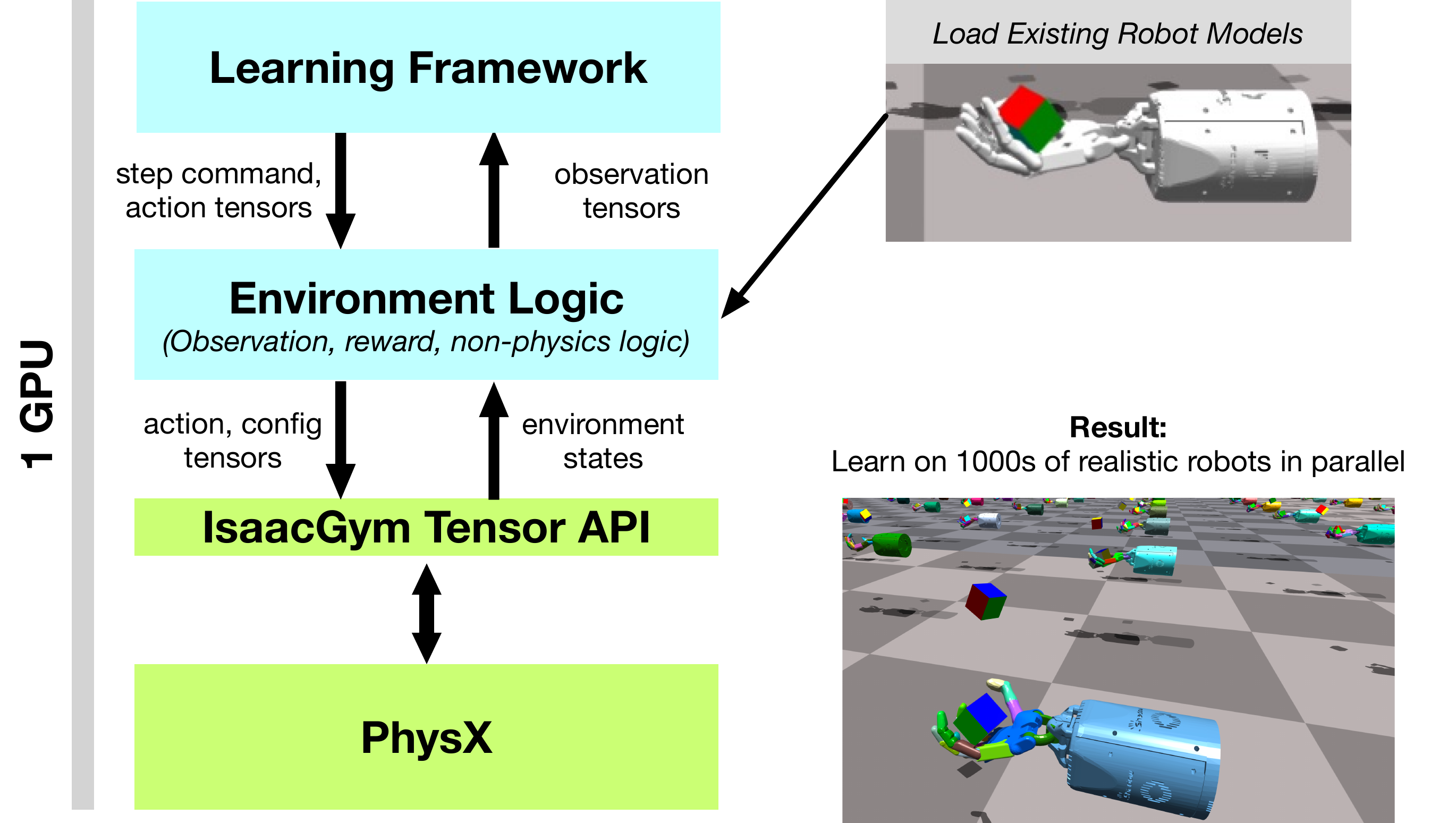}
\caption{An illustration of the Isaac Gym pipeline. The Tensor API provides an interface to Python code to step the PhysX backend, as well as get and set simulator states, directly on the GPU, allowing a 100-1000x speedup in the overall RL training pipeline while providing high-fidelity simulation and the ability to interface with existing robot models.}
\label{fig:system}
\end{figure}

Isaac Gym provides a straightforward API for creating and populating a scene with robots and objects, supporting loading data from the common URDF and MJCF file formats. Each environment is duplicated as many times as needed, while preserving the ability for variations between copies (\textit{e.g.} via Domain Randomization \citep{openai-dr}). Environments are simulated simultaneously in parallel without interaction with other environments. Using a fully GPU-accelerated simulation and training pipeline can help lower the barrier for research, enabling solving of tasks with a single GPU that were previously only possible on massive CPU clusters. Isaac Gym also includes a basic Proximal Policy Optimization (PPO) implementation and a straightforward RL task system, but users may substitute alternative task systems or RL algorithms as desired. While the included examples use PyTorch, users should also be able to integrate with TensorFlow training libraries with further customization. An overview of the system is provided in Figure~\ref{fig:system}.

\newpage

\begin{mdframed}[
linecolor=black!40,
outerlinewidth=1pt,
roundcorner=.5em,
innertopmargin=1ex,
innerbottommargin=.5\baselineskip,
innerrightmargin=1em,
innerleftmargin=1em,
shadow=true,
shadowsize=6,
shadowcolor=black!20,
frametitle={Summary of Results},
frametitlebackgroundcolor=backcolour,
frametitlerulewidth=10pt%
]

Our major contributions include:
\begin{itemize}
   \item Development of high-fidelity GPU-accelerated robotics simulator for robot learning tasks.
   \item A Tensor API in Python providing direct access to physics buffers by wrapping them into PyTorch tensors without going through any CPU bottlenecks. 
   \item Implementation of multiple highly complex robotic manipulation environments which can be simulated at hundreds of thousands of steps per second on a single GPU.
   \item High-performance training results using Isaac Gym with Deep Reinforcement Learning on challenging robotic environments.
\end{itemize}
Our major empirical results include:
\begin{itemize}
    \item We achieve significant speed-ups in training various simulated environments: Ant and Humanoid environments can achieve performant locomotion in 20 seconds and 4 minutes respectively, ANYmal \cite{ANYmal:etal:2016} in under 2 minutes, Humanoid character animation using AMP \cite{2021-TOG-AMP} in 6 minutes and cube rotation with Shadow Hand in 35 minutes all on a \textbf{single NVIDIA A100 GPU}.
    
    \item Additionally, we reproduce OpenAI Shadow Hand cube training setup \cite{openai-sh} with asymmetric actor-critic and domain randomization. We show that we can achieve similar performance to OpenAI results of 20 consecutive successes with feed forward and 37 consecutive successes with LSTM networks with a success tolerance of 0.4 rad in about 1 hour and 6 hours on an average respectively on A100. In contrast, OpenAI effort required 30 hours and 17 hours respectively on a combination of a CPU cluster (384 CPUs with 16 cores each) and 8 NVIDIA V100 GPUs with MuJoCo \cite{MuJoCo:etal:2012} using a conventional RL training setup. 
    It is worth mentioning that since OpenAI show results with only 1 seed, comparing our best seed we find that we achieve 37 consecutive successes with LSTMs in just 2.5 hours.
    
    \item We also demonstrate sim-to-real transfer results on ANYmal and TriFinger which further showcases the ability of our simulator to perform high-fidelity contact rich manipulation.
\end{itemize}
\end{mdframed}

\section{Background}

\subsection{Parallelization Strategy}

There are many approaches to parallelizing physics simulations.  We outline these approaches here and justify our design decisions in the context of GPU-accelerated simulation tailored towards learning algorithms.  Isaac Gym was developed to maximize the throughput of physics-based machine learning algorithms with particular emphasis on simulations that require large numbers of environment instances executing in parallel.

\subsubsection{CPU Simulations}

When physics simulation runs on CPU, multiple threads can be used to distribute computation among the available cores.  The most straightforward strategy is simulating one environment instance per thread.  In this approach, scaling is limited by the number of physical cores in the system.  On a 64-core hyper-threaded CPU, we could run up to 128 environments in parallel, but CPUs with a large number of cores are typically clocked lower to prevent overheating.  Running tens or hundreds of threads comes with other potential pitfalls including synchronization, context-switching overhead, and memory bandwidth limitations.  To scale further, we would need to use a multi-CPU setup or build a cluster, which introduces additional communication overhead.

Running a single environment instance per thread in its own dedicated physics scene can be inefficient.  There is some overhead involved in setting up, executing, and gathering the results of each physics step.  The simpler the environment, the more significant the overhead.  To mitigate this, we can pack multiple environments into a single physics scene.  For example, we could split 1024 environments into eight physics scenes with 128 environments each.  Each scene can run in its own thread.  Extra provisions are needed to ensure that environments in the same scene do not interact with each other physically, which can be done using contact filtering and other methods.

\subsubsection{GPU Simulations}

Running the physics simulation on GPU can result in significant speedups, especially for large scenes with thousands of individual actors.  On the GPU, the physics engine can parallelize computations at the level of individual shapes, bodies, or joints.  High-end GPUs require many thousands of objects to effectively utilize their streaming multiprocessor architecture.  This makes them a good match for running simulations with thousands of environment instances.  On GPU, we don't need to worry about splitting the environments into multiple scenes.  In fact, the opposite is generally true - we want to pack everything into a single scene to take advantage of the deep fine-grained parallelism and maximize the overall throughput. 

Physics simulations on a GPU is not new.  In previous work, we demonstrated good results with running GPU physics simulations for reinforcement learning \cite{Liang:etal:CoRL2018}. In this work, the GPU was used as a co-processor that accelerates the physics simulation, while the API for getting physics state and applying controls was CPU-based.  There are, however, performance bottlenecks with this strategy.  In a reinforcement learning pipeline, physics simulation is just one part of the system.  After a physics step, we need to get the latest physics state to compute observations and rewards.  If these computations are done on the CPU, we need to transfer the physics state from the GPU.  While modern hardware architectures can achieve impressive data transfer speeds, large simulations can incur nontrivial overhead.  Then, the raw physics state needs to be processed on the CPU to compute observations and rewards, which is subject to similar parallelization challenges as discussed above due to the limited number of CPU cores. Next, the observations and rewards need to be copied from system memory back to device memory for the reinforcement learning algorithm.  After the learning step, a set of actions is generated by the policy network on the GPU.  These actions need to be copied to the CPU so that they can be converted to physics simulation inputs.  Those inputs end up being copied to the GPU again to run the next step of physics simulation on the device. 

\begin{figure}[h]
\centerline{
\includegraphics[width=0.67\textwidth]{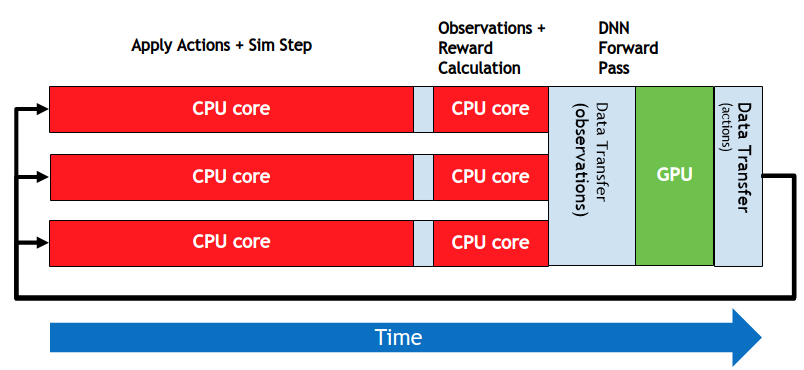} 
\hfill
\includegraphics[width=0.33\textwidth]{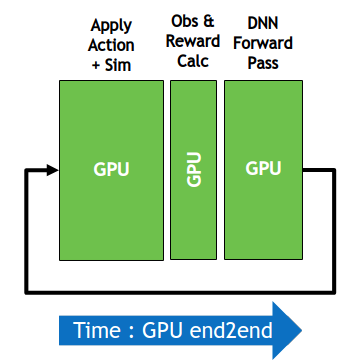}
}
\centerline{
\makebox[0.66\linewidth][c] { \footnotesize{(a) Traditional RL experience collection.} }
\hfill
\makebox[0.33\linewidth][c] { \footnotesize{(b) Isaac Gym experience collection.} }
}
\caption{\small
\textbf{(a)} Traditional RL experience collection pipelines often use CPU based physics engines which quickly become the bottleneck. \textbf{(b)} In contrast, Isaac Gym not only runs physics on the GPU but also directly copies the physics data to the deep neural network framework using CUDA interoperatability without ever using CPU in the process. This massively improves the performance of RL training process leading to significantly faster training times.} 
\label{fig:traditional_vs_end2end}
\end{figure}

Isaac Gym eliminates those inefficiencies by keeping all of the computations on the GPU.  Stepping physics, computing observations and rewards, and applying actions are performed on the GPU without ever copying large quantities of data between devices.  Two new features were added to PhysX to facilitate this.  First, PhysX GPU simulations can run without fetching the results to the CPU after every step.  Second, a new direct GPU API was added to access the current state, submit state changes, and apply control inputs in GPU buffers. In Figure \ref{fig:traditional_vs_end2end}, we contrast the traditional RL experience collection pipeline with our high throughput fully GPU-based pipeline. 

\subsection{Simulation Setup}

Isaac Gym provides a simple procedural API to create environments and populate them with actors.  It supports loading assets from URDF and MJCF file formats.  These assets can be instanced multiple times in simulation environments to create actors.  In the underlying PhysX engine, single-body actors are created as rigid dynamics and multi-body actors are created as reduced coordinate articulations.  During the setup phase, users can set initial actor poses, configure joint drives, and customize rigid body properties and physics materials.  Most joint and rigid body properties can be changed during the simulation as well, which facilitates domain randomization without stopping and restarting the simulation. Below we provide definitions of some useful terms.

\textbf{Actor:} An entity composed of rigid bodies connected via joints. It can be created via direct loading of a URDF model or XML file composed of either meshes or primitive shapes.

\textbf{Rigid Bodies:} A primitive shape or a mesh model that comprises an actor is called a rigid body. The positions, rotations and velocities of a rigid body can be obtained via the API.

\textbf{DOF States:} Rigid bodies are connected by various joints. A joint can have 0 or more degrees of freedom.  Fixed joints have no DOFs, revolute and prismatic joints have 1 DOF and spherical joints have 3 DOFs. The DOF states, which include joint position and velocity, can be obtained via the API.

The setup code runs on the CPU to allow flexibility in per-instance setup, but once the simulation starts Isaac Gym provides a tensor API that can be used to interact with the running simulation on either CPU or GPU.  Users can specify the device to be used for the simulation and the tensor interface in the simulation parameters.

\begin{figure}[h]
  \centering
  \includegraphics[width=\textwidth]{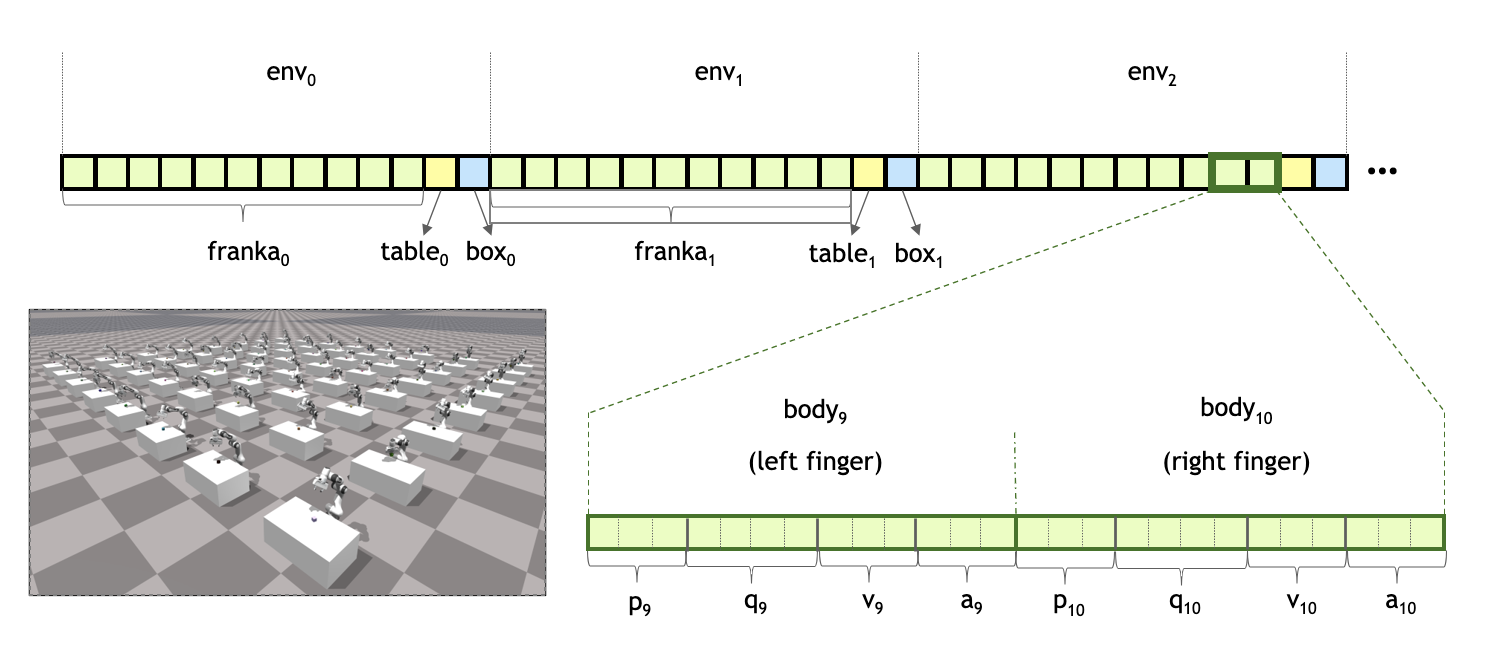}
\caption{\small Tensors associated with the scene composed of multiple copies of the same environment simulating different variations all running in parallel. Each actor (\textit{e.g.} table, box or franka) has various bodies and their corresponding positions, quaternions and velocities are stored directly in PyTorch tensors.}
\label{fig:tensor_api}
\end{figure}

\subsection{Tensor API}

Isaac Gym provides a data abstraction layer over the physics engine.  This allows us to support multiple physics engines with a shared front-end API.  In this work, the physics engine is PhysX, although some limited tensor API functionality is available with the FleX physics engine as well.

Instead of calling physics engine functions directly, users can access all of the physics data in flat buffers. This data-oriented approach allows us to eliminate a lot of overhead caused by looping over tens of thousands of individual simulation actors in user code.  Physics state is exposed to Python users as global tensors.  For example, all rigid body states can be found in a single rigid body state tensor. Figure \ref{fig:tensor_api} shows a typical Isaac Gym scene composed of various copies of the same environment simulating different variations all running in parallel and the corresponding tensors associated with it. Control inputs can be applied using global tensors as well.  For example, applying forces to all rigid bodies in the simulation can be done using a single function call that takes a tensor containing all of the forces. Users can create custom views or slices of the global tensors to suit their needs.  When multiple environment instances are packed into the simulation, it is possible to create custom views of the data with the environment index as one of the dimensions.  This makes it easy to vectorize observation and reward computations by running GPU kernels on multiple environments in parallel.

\subsubsection{Python Interface}

The core of Isaac Gym is implemented using C++ and CUDA.  It is completely independent of any Python frameworks commonly used in machine learning.  To make the data easily accessible to Python users, Isaac Gym provides utilities that can "wrap" the raw data buffers as tensor objects in common machine learning frameworks like PyTorch. %
The tensor-wrapping utilities make it possible to share the native CPU or GPU buffers with Python without any copying overhead.

A powerful feature of Isaac Gym is the ability to run the same code on either CPU or GPU by simply toggling a flag.  Python users do not need to write custom CUDA or C++ kernels to compute observations, rewards, or actions.  When physics state and control tensors are wrapped as PyTorch tensors, users can take advantage of TorchScript JIT to compile their Python functions to lower level scripts which orchestrate the training pipeline quickly.

\subsubsection{Physics State Tensors}

\begin{table}[t]
\centering
\begin{tabular}{| p{0.2\textwidth} | p{0.5\textwidth} | p{0.09\textwidth} | p{0.08\textwidth} |}
\hline
\rowcolor{Light} Tensor & Description     & Shape    & Usage                    \\
\hline
Actor root state & State of all actor root bodies (position, orientation, linear and angular velocity).               & $(N_A, 13)$  & Get/Set \\
\hline
DOF state  & State of all degrees of freedom (position and velocity). & $(N_D, 2)$  & Get/Set \\
\hline
Rigid body state & State of all rigid bodies (position, orientation, linear and angular velocity).               & $(N_B, 13)$  & Get \\
\hline
DOF forces & Net forces experienced at each degree of freedom. & $N_D$  & Get \\
\hline
Rigid body forces & Rigid body forces and torques experienced at force sensor locations. & $(N_F, 6)$  & Get \\
\hline
Net contact forces & Net forces experienced by each rigid body. & $(N_B, 3)$  & Get \\
\hline
Jacobian matrix & Jacobian matrices for a homogeneous group of actors. & Variable  & Get \\
\hline
Mass matrix & Generalized mass matrices for a homogeneous group of actors. & Variable  & Get \\
\hline
\end{tabular}
\vspace{2mm}
\caption{Physics state tensors.  $N_A$ is the total number of actors, $N_B$ is the total number of rigid bodies (including articulation links), $N_D$ is the total number of degrees of freedom, and $N_F$ is the total number of rigid body force sensors.}
\label{tensors:state:table}
\end{table}

Physics state tensors are used to obtain state snapshots of a running simulation.  Isaac Gym allows for interacting with the simulation using maximal and reduced coordinates.  Physics state includes the kinematic state of rigid bodies and degrees of freedom (DOFs).  Rigid body state consists of position, orientation (quaternion), linear velocity, and angular velocity.  DOF state includes position and velocity. In the code snippet below we show how to access them through the API.

\begin{lstlisting}[language=Python, title={Obtaining state information by wrapping physics buffers into PyTorch tensors. CUDA interoperability allows copying the data directly without ever going through the host.} ,label={lst:gym_tensors}, captionpos=b]
# Acquire tensor descriptors 
# - Raw storage buffer independent of client framework.
# - Storage will be on GPU if using GPU pipeline, CPU otherwise.
# - Same code for CPU and GPU just different device.

root_state_desc = gym.acquire_actor_root_state_tensor(sim)
dof_state_desc  = gym.acquire_dof_state_tensor(sim)

# PyTorch interop
# No data copying, just wrap the gym buffers as torch tensors.

# The root state tensor captures the state of the root bodies of all actors. 

root_states = gymtorch.wrap_tensor(root_state_desc)
dof_states  = gymtorch.wrap_tensor(dof_state_desc)

# obtaining physics states 
# Physics state includes kinematic states of rigid bodies and degrees of freedom (DOFs).

root_state_vec = root_states.view(num_envs, actors_per_env, 13)
dof_state_vec  = dof_states.view(num_envs, dofs_per_env, 13)

root_p = root_states[..., 0:3]   # positions of rigid bodies
root_q = root_states[..., 3:7]   # rotations, in quaternions, of rigid bodies
root_v = root_states[..., 7:10]  # linear velocities of rigid bodies
root_a = root_states[..., 10:13] # angular velocities of rigid bodies 

dof_p  = dof_state_vec[..., 0]   # joint positions
dof_v  = dof_state_vec[..., 1]   # joint velocities
\end{lstlisting}

Revolute DOFs use radians and linear DOFs use meters for units.  Additional state data includes contact forces, rigid body force sensors, and DOF force sensors.  To support operational space control and inverse kinematics applications, Isaac Gym also provides Jacobian and generalized mass matrices which can be obtained for articulated actors. 

The available state tensors are listed in Table ~\ref{tensors:state:table}.  Most of the state tensors are read-only, except the root state tensor and the DOF state tensor.  These two tensors play a special role, because they can be used to fully set the poses and velocities of actors.  This can be used during environment resets, when new poses are generated or original poses need to be restored.  The root state tensor captures the state of the root bodies of all actors.  For single-body actors, the root state fully captures their poses and velocities in maximal coordinates.  For articulated actors, the root state can be used to "teleport" them without changing the poses of the descendant articulation links.  The DOF state tensor can be used to configure the descendant articulation links using reduced coordinates.  Setting new DOF states does not affect the root state.  For fixed-base articulated actors, such as mounted robotic arms, the DOF state tensor fully captures the articulation poses and velocities. Users can apply new root and DOF states for all actors at once or to a limited subset using an index buffer.  This allows resetting a subset of environments without affecting the rest.

\subsubsection{Physics Control Tensors}

Physics simulation inputs include forces, torques, and PD controls such as position and velocity targets.  Forces and torques can be applied to rigid bodies and DOFs.  PD targets are applied to DOFs that have been configured to use position or velocity drives.  Users can configure the drive parameters like stiffness and damping using a separate API.  Table ~\ref{tensors:control:table} lists the available control tensors.  The control tensors are typically created in a higher-level framework like PyTorch, but can be efficiently shared with Isaac Gym using the tensor-wrapping utilities.

\begin{table}[H]
\centering
\begin{tabular}{| p{0.22\textwidth} | p{0.3\textwidth} | p{0.09\textwidth} | p{0.27\textwidth} |}
\hline
\rowcolor{Light} Tensor & Description     & Shape    & Applied to                    \\
\hline
DOF actuation forces                     & \begin{tabular}[c]{@{}l@{}}Torques or linear forces to be \\ applied to degrees of freedom.\end{tabular} & $N_D$      & All actors or indexed subset \\ \hline
DOF position targets                     & \begin{tabular}[c]{@{}l@{}}PD position targets for \\ degrees of freedom.\end{tabular}                   & $N_D$      & All actors or indexed subset \\ \hline
DOF velocity targets                     & \begin{tabular}[c]{@{}l@{}}PD velocity targets for \\ degrees of freedom.\end{tabular}                   & $N_D$      & All actors or indexed subset \\ \hline
Rigid body forces                        & \begin{tabular}[c]{@{}l@{}}Forces to be applied to \\ rigid bodies.\end{tabular}                         & $(N_B, 3)$ & All rigid bodies             \\ \hline
Rigid body torques                       & \begin{tabular}[c]{@{}l@{}}Torques to be applied to \\ rigid bodies.\end{tabular}                        & $(N_B, 3)$ & All rigid bodies             \\ \hline
\end{tabular}
\vspace{2mm}
\caption{Physics control tensors.  $N_B$ is the total number of rigid bodies (including articulation links) and $N_D$ is the total number of degrees of freedom.}
\label{tensors:control:table}
\end{table}

\section{Physics Simulation}

Robots are simulated using PhysX \cite{nvidia-physx} reduced coordinate articulations. Any individual rigid bodies may be simulated using either maximal coordinate rigid bodies or single-link reduced coordinate articulations. Articulations with a single link and rigid bodies are equivalent and interchangeable. We also support tendons to actuate degrees of freedom and they are simulated in PhysX using Fixed Tendon mechanics. The physics of tendons are described in detail in Section~\ref{sec:tendon-details}. We tested the dynamics of tendons using the Shadow Hand simulation environment, described in Section~\ref{sec:env:shadow-hand}.

We use the Temporal Gauss Seidel (TGS) ~\cite{TGS:2019} solver to compute the future states of objects in our physics simulation. The TGS solver uses the observation that sub-stepping a simulation with a single gauss-seidel solver iteration yields significantly faster convergence than running larger steps with more solver iterations. It folds this process efficiently into the iteration process, calculating the velocity at the end of each iteration and accumulating these velocities (scaled by $dt/N$, where $N$ is the number of iterations) into a per-body accumulated delta buffer. This delta buffer is projected onto the constraint Jacobians and added to the bias terms in the constraints. This approach adds only a few additional operations to a more traditional Gauss-Seidel solver, producing almost identical performance cost per-iteration. However, it achieves the same effect on convergence as having sub-stepped the simulation without the computational expense. With positional joint constraints, an additional rotational term is calculated for joint anchors to improve handling of non-linear motion to avoid linearization artifacts. This term is not necessary (and in fact undesirable) to add to contacts. Various parameters exposed to the user to tune the simulator are described in Table \ref{sim:params:table}.

\begin{table}[h]
\centering
\begin{tabular}{l|l}
\rowcolor{Light} Parameter               & Description          
\\ \hline
Delta time (dt)                      & Controls time-step size  \\
Gravity                 & Controls the gravity in the scene                 \\
Collision filtering     & Filters collisions between shapes      \\
Position iterations     & Biased (velocity + positional error correcting) solver iterations                        \\
Velocity iterations     & Unbiased (velocity error only correcting) solver iterations                         \\
Max bias coefficient    & Limits the magnitude of position error bias
Friction                        \\
Restitution             & Controls bounce     \\
Static/dynamic friction & Static and dynamic friction coefficients  \\
Bounce threshold        & Relative normal velocity limit below which restitution is ignored         \\ \hline
Rest offset            & \begin{tabular}[c]{@{}l@{}}Distance at which shapes are held separated. Default is 0 but can \\ be increased to hold objects at gap. Useful for thin objects.\end{tabular}  \\ \hline
Friction offset threshold & \begin{tabular}[c]{@{}l@{}} Distance at which friction anchors are discarded \\ (static friction depends on friction anchor caching)\end{tabular} \\ \hline
Solver offset slop & \begin{tabular}[c]{@{}l@{}} An epsilon value used to correct for round-off errors in contact \\ gen. Corrects small skew effects with rolling spheres or capsules. \end{tabular} \\ \hline
Friction correlation distance & \begin{tabular}[c]{@{}l@{}} Distance at which contacts are merged into a single \\ friction constraint \end{tabular}\\ \hline
Max force & Per-body and per-contact force limits \\ 
Drive stiffness & Positional error correction coefficient of a PD controller \\ 
Drive damping & Velocity error correction coefficient of a PD controller \\ 
Joint friction & Per-joint frictional term. Simulates dry friction in a joint.\\
Joint armature & Per-joint armature term - simulates motor inertia. \\ 
Body/link Damping & World-space linear/angular damping on each body/link \\
Max velocity & Linear/angular velocity limits per-body \\ \hline
\end{tabular}
\vspace{2mm}
\caption{Parameters exposed to tune the simulator.}
\label{sim:params:table}
\end{table}

\section{Environments}
\label{sec:envs-used}
We implemented a diverse set of environments covering different application areas. Here we describe a subset of representative examples and key points related to the training. Benchmark results on the simulation performance and training results are presented in the subsequent sections. 

All environments are trained using the Proximal Policy Optimization algorithm \citep{schulman2017proximal}, using rl\_games, a highly-optimized GPU end-to-end implementation from \citep{rl-games}. This implementation vectorizes observations and actions on GPU allowing us to take advantage of the parallelization provided by the simulator. We list the environments used in our experiments below:

\begin{mdframed}[
linecolor=black!40,
outerlinewidth=1pt,
roundcorner=.5em,
innertopmargin=1ex,
innerbottommargin=.5\baselineskip,
innerrightmargin=1em,
innerleftmargin=1em,
shadow=true,
shadowsize=6,
shadowcolor=black!20,
frametitle={Environments used},
frametitlebackgroundcolor=backcolour,
frametitlerulewidth=10pt
]

\begin{enumerate}
    \item \textbf{Locomotion Environments}
    \begin{itemize}
        \item Ant
        \item Humanoid
        \item Ingenuity
        \item ANYmal 
    \end{itemize}
    \item \textbf{Franka Cube Stacking}
    \item \textbf{Humanoid Character Animation}
    \item \textbf{Robotic Hands}
    \begin{itemize}
        \item Shadow
        \item Allegro
        \item Trifinger 
    \end{itemize}
\end{enumerate}
\end{mdframed}

While Ant and Humanoid are relatively simple environments popularised by MuJoCo continuous control benchmarks, the strength of our simulator really shines when training on environments that are rich in complexity particularly robotic hands. Various meta-data related to simulation setup for these environments is in Table \ref{sim-set-up-envs}.

\paragraph{Key Experimental Details}
\begin{itemize}
\item Unless stated otherwise, all experiments are done on a system with a \textbf{single NVIDIA A100 GPU} and a single 3.7GHz Intel i7-8700K CPU

\item All training runs for each environment are \textbf{averaged over 5 seeds}. The reward curves are plotted with $\mu \pm \sigma$ regions. 

\item All the environments by default follow symmetric actor-critic approach with shared observations as well as shared network for policy and value functions. Sharing the network allows faster forward passes and improves training. 

\item Moreover, for Shadow Hand and TriFinger, we also use an asymmetric actor critic approach \citep{asymmetric-ac} with policy observations that are closest to real world settings while value function receives privileged state information from simulation as well as the observations received by the policy. This approach is naturally suited for sim-to-real transfers.

\item For all environments trained with feed forward networks we use a discount factor of $\gamma=0.99$ while LSTM networks use $\gamma=0.998$. We use a GAE discount factor, $\lambda=0.95$ and clipping $\epsilon=0.2$. Also, we use an adaptive learning rate and varying KL thresholds per environment. 

\item Detailed hyper-parameters for each training task are shown in Table~\ref{tab:hyperparameters}. Rewards and observations for each environment we used can be found in Appendix \ref{sec:env-details}. 
\end{itemize}

\begin{table}[h]
\centering
\begin{tabular}{| p{0.23\textwidth} | p{0.26\textwidth} | p{0.14\textwidth} | p{0.11\textwidth} | p{0.13\textwidth} |}
\hline
\rowcolor{Light} Environment & Control Type     & Simulation $dt$ & Control $dt$  & Action Dims                    \\
\hline
Ant                     & Joint Torques & 1/120 s  & 1/60 s    & 8 \\ \hline
Humanoid                     & Joint Torques & 1/120 s  & 1/60 s    & 21 \\ \hline
Ingenuity                     & Rigid Body Forces                   & 1/200 s    & 1/100 s   & 6 \\ \hline
ANYmal                        & Joint Position Targets                         & 1/200 s  & 1/50 s & 12          \\ \hline
Franka Cube Stacking                       & Operation Space Control                       & 1/60 s & 1/60 s & 7             \\ \hline
Shadow Hand Standard                     & Joint Position Targets                       & 1/120 s & 1/60 s & 20             \\ \hline
Shadow Hand OpenAI                      & Joint Position Targets                       & 1/120 s & 1/20 s & 20             \\ \hline
Allegro Hand                       & Joint Position Targets                       & 1/120 s & 1/20 s & 16            \\ \hline
TriFinger                       & Joint Torques                       & 1/200 s & 1/50 s & 9             \\ \hline
\end{tabular}
\vspace{2mm}
\caption{Simulation setup for the environments.}
\label{sim-set-up-envs}
\end{table}

\section{Characterising Simulation Performance}

We first characterise the simulation performance as a function of number of environments. As we vary this number, we aim to keep the overall experience an RL agent observes constant by decreasing the horizon length proportionally (\textit{i.e.} number of steps in PPO) for a fair comparison. While we provide detailed training studies for many environments later, we characterise simulation performance only for \textbf{Ant}, \textbf{Humanoid} and \textbf{Shadow Hand} as they are sufficiently complex to test the limits of the simulation and also represent a gradual increase in the complexity. All three environments use feed forward networks for training. 

\subsection{Ant}

\begin{figure}[h]
  \centerline{
  \includegraphics[width=.41\textwidth]{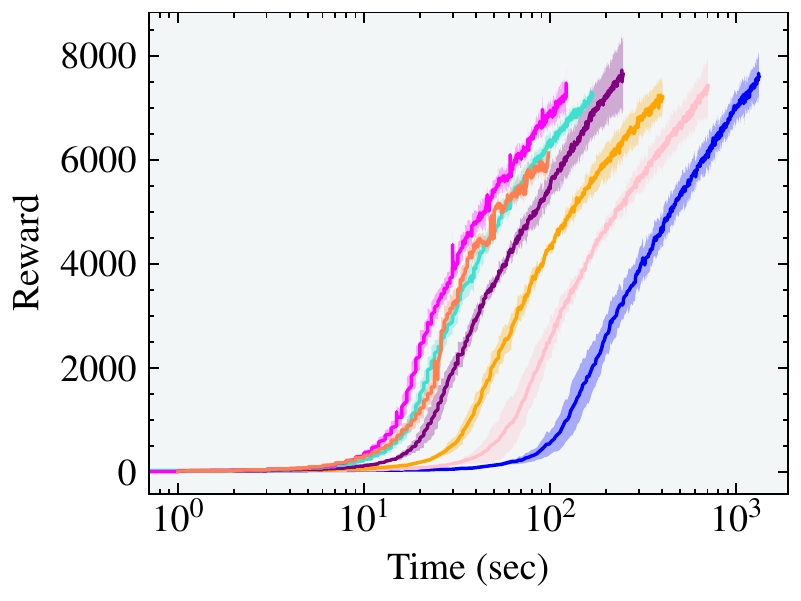}
  \includegraphics[width=.62\textwidth]{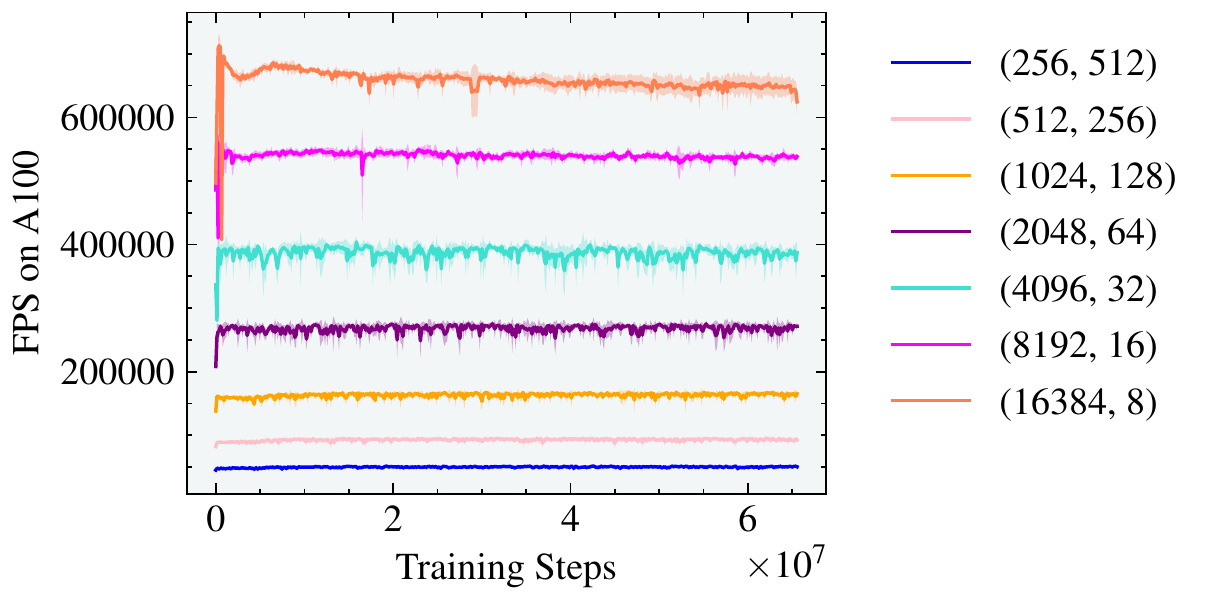}
  }
  \centerline{
    \makebox[0.5\linewidth][c] { \footnotesize{(a) Rewards} }
    \hfill
    \makebox[0.5\linewidth][l] { \footnotesize{(b) Total number of environment steps per second} }
  }
\caption{Rewards and effective FPS with respect to number of parallel environments for the Ant experiment. Best training time is achieved with 8192 environments and a horizon lengths of 16.}
\label{fig:ant_wrt_envs}
\end{figure}

We first experiment with the standard Ant environment where the agent is trained to run on a flat ground. We find that as the number of agents is increased, the training time, as expected, is reduced \textit{i.e.} changing the number of environments from 256 to 8192 --- an increase by 5 orders of magnitude --- leads to a reduction in training time to reach 7000 reward by an order of magnitude from 1000 seconds (\textasciitilde{16.6} minutes) to 100 seconds (\textasciitilde{1.6} minutes). \textbf{However, note that Ant reaches performant locomotion at 3000 reward in just 20 seconds on a single GPU.}

Since Ant is one of the simplest environments to simulate, the number of parallel environment steps per second as depicted in the Figure \ref{fig:ant_wrt_envs}(b)
can go as high as 700K. We do not observe gains when increasing the number of environments from 8192 to 16384 due to reduced horizon length. 

\subsection{Humanoid}
\begin{figure}[h]
\vspace{-3mm}
\centerline{
\includegraphics[width=0.41\textwidth]{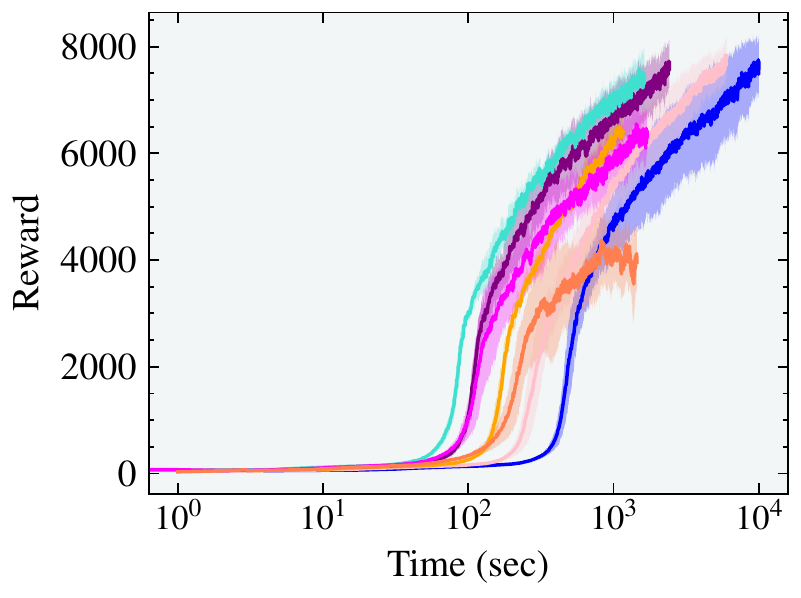} 
\hfill
\includegraphics[width=0.62\textwidth]{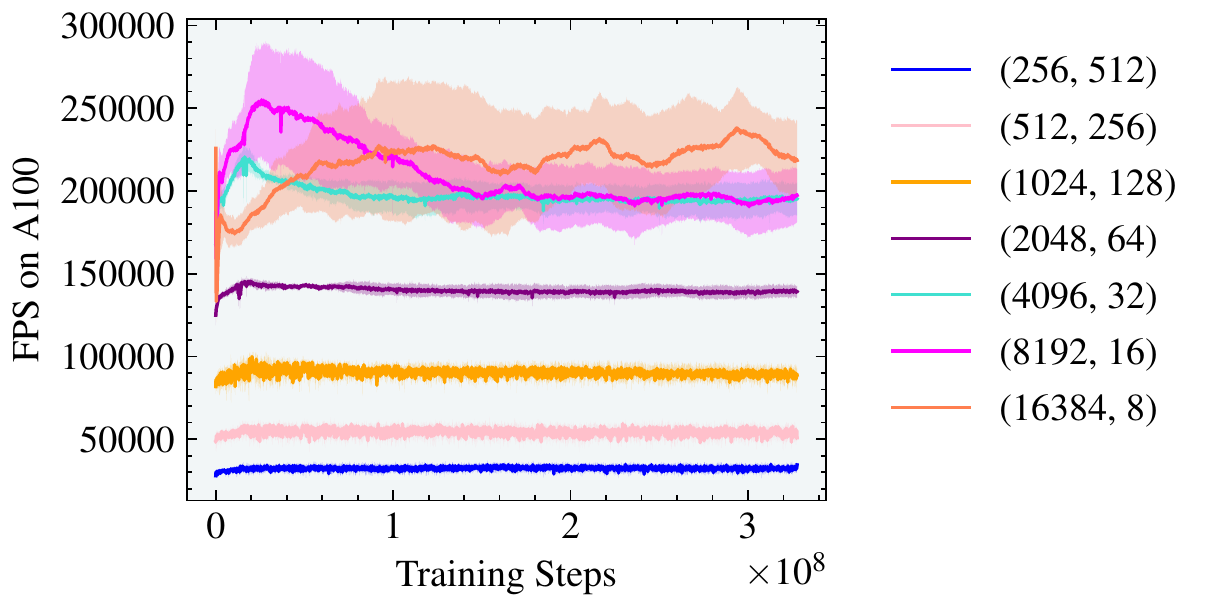} 
}
\centerline{
\makebox[0.5\linewidth][c] { \footnotesize{(a) Rewards} }
\hfill
\makebox[0.5\linewidth][l] { \footnotesize{(b) Total number of environment steps per second} }
}
\caption{\small Rewards and effective FPS with respect to number of parallel environments for the Humanoid experiment. Best training time is achieved with 4096 environments and a horizon lengths of 32.} 
\label{fig:humanoid-wrt-envs}
\end{figure}

\begin{figure}[h]
\vspace{-3mm}
\centerline{
\includegraphics[width=0.41\textwidth]{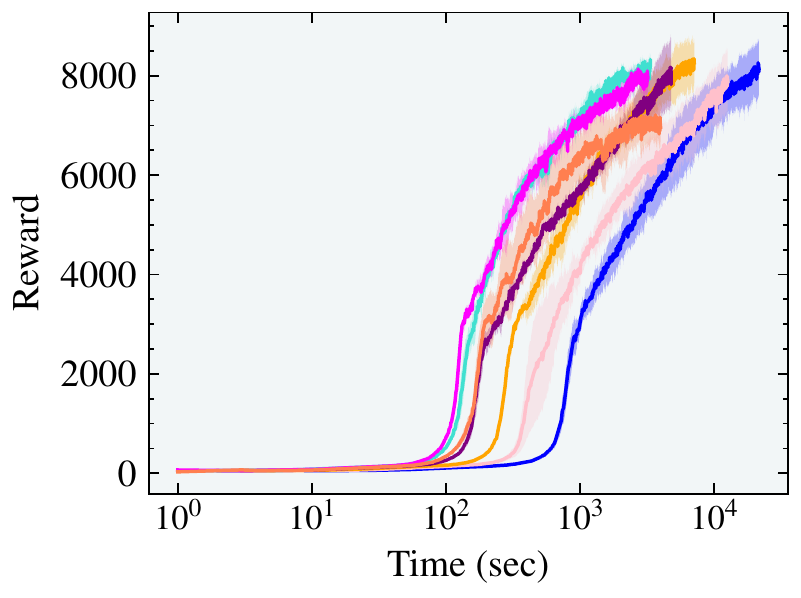}
\hfill
\includegraphics[width=0.62\textwidth]{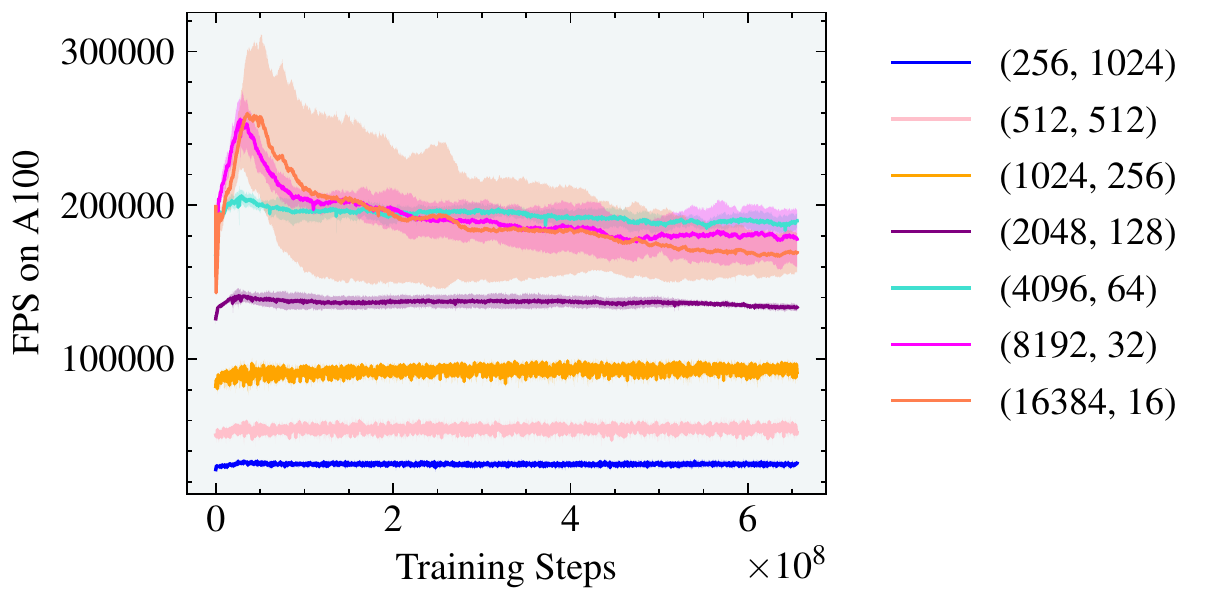}
}
\centerline{
\makebox[0.5\linewidth][c] { \footnotesize{(a) Rewards} }
\hfill
\makebox[0.5\linewidth][l] { \footnotesize{(b) Total number of environment steps per second} }
}
\caption{\small Rewards and effective FPS with respect to number of parallel environments for the Humanoid experiment. Best training time is achieved with both 4096 and 8192 environments and horizon lengths of 64 and 32 respectively.} 
\label{fig:humanoid-wrt-envs-2}
\end{figure}

The Humanoid environment has more degrees of freedom and requires the agent to discover the gait that lets itself balance on two feet and walk on the ground. As observed in Figure \ref{fig:humanoid-wrt-envs} and Figure \ref{fig:humanoid-wrt-envs-2}, the training times are increased by an order of magnitude compared to the Ant in Figure \ref{fig:ant_wrt_envs}.

We also note in Figure \ref{fig:humanoid-wrt-envs} that as the number of agents is increased, in this case, from 256 to 4096, the training time needed to reach the highest reward of 7000 is reduced by an order of magnitude from $10^4$ seconds (\textasciitilde{2.7} hours) to $10^3$ seconds (\textasciitilde{17} minutes). \textbf{However, performant locomotion starts happening at around a reward of 5000 at a training time of just 4 minutes.} Going beyond 4096 environments for this set up resulted in no further gains and in fact led to both increase in training time and sub-optimal gaits. We attribute this to the complexity of the environment that makes it challenging to learn walking at such small horizon lengths.  

We verified this by training on another set of environment and horizon length combinations where horizon length was increased by a factor of 2 compared to Figure \ref{fig:humanoid-wrt-envs}. As shown in the Figure \ref{fig:humanoid-wrt-envs-2}, the humanoid is able to walk even with 8192 and 16384 environments which have small horizon lengths of 32 and 16 respectively but sufficiently long to enable learning.

Also worth noting that due to the increased degrees of freedom the number of parallel environment steps per second is reduced from 700K for Ant to 200K for Humanoid as shown in Figures \ref{fig:humanoid-wrt-envs} and \ref{fig:humanoid-wrt-envs-2}.

\subsection{Shadow Hand}

\begin{figure}[h]
\vspace{-3mm}
\centerline{
\includegraphics[width=0.41\textwidth]{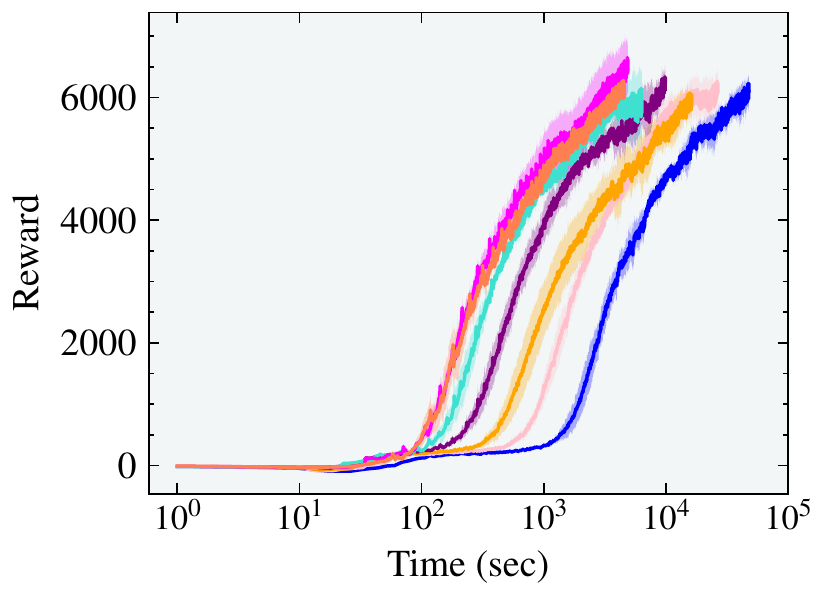} 
\hfill
\includegraphics[width=0.60\textwidth]{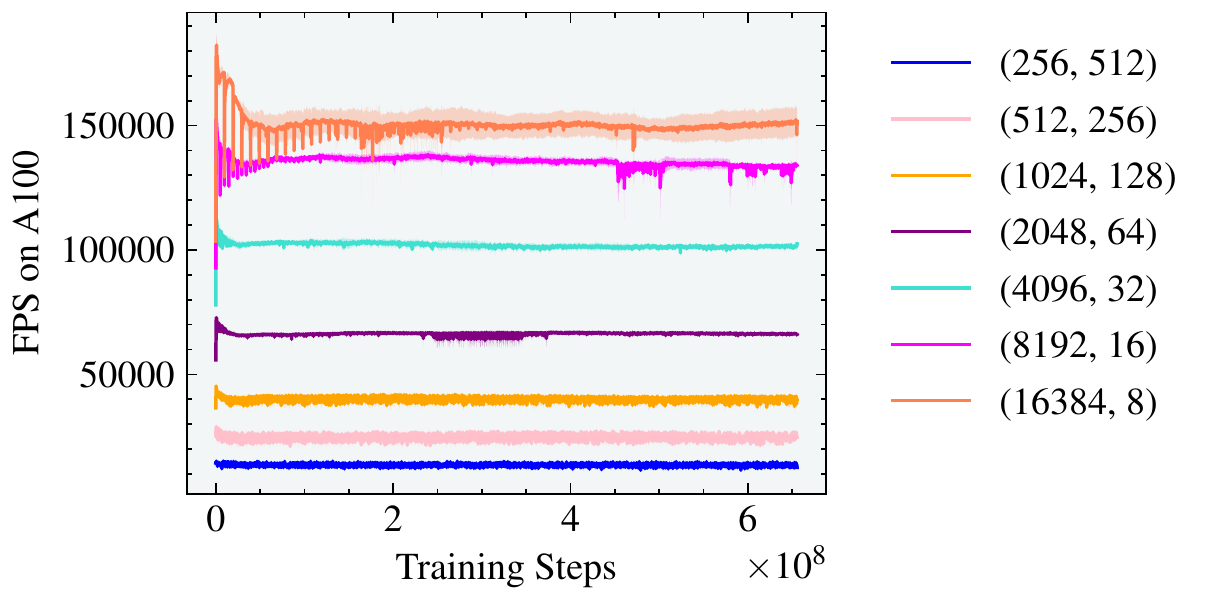}
}
\centerline{
\makebox[0.5\linewidth][c] { \footnotesize{(a) Rewards.} }
\hfill
\makebox[0.5\linewidth][l] { \footnotesize{(b) Total number of environment steps per second} }
}
\caption{\small Rewards and effective FPS with respect to number of parallel environments for the Shadow Hand experiment. Best training time is achieved with both 8192 and 16384 environments and horizon lengths of 16 and 8 respectively.} 
\label{fig:shadowhand-wrt-envs}
\end{figure}

Lastly, we experiment with Shadow Hand \cite{openai-sh} to learn to rotate a cube resting on the palm to a target orientation using the fingers and the wrist. This task is challenging due to the number of DoFs involved and the contacts that are made and broken during the process of rotation. Our results with Shadow Hand environment follow similar trends. As the number of agents is increased, in this case, from 256 to 16384, the training time is reduced by an order of magnitude from $5\times10^4$ seconds (\textasciitilde{14} hours) to $3\times10^3$ seconds (\textasciitilde{1} hour). \textbf{We find that the environment reaches performant dexterity of 10 consecutive successes at reward of 3000 in just 5 minutes.}\footnote{The experiments used Shadow Hand Standard variant as explained in Section \ref{sec:env:shadow-hand}.} Further performance improvements continue to happen as more experience is collected. Additionally, we find that the horizon length of 8 for 16384 agents still allows learning re-posing the cube. The maximum effective frame-rate of 150K number of parallel environment steps per second was achieved with 16384 agents. 

\section{Characterising Environment Performance}

We now provide details and performance metrics for individual environments mentioned in Section \ref{sec:envs-used} trained using a PPO implementation that operates on vectorised states and actions.

\begin{figure}[t]
\centering
\hspace{-1.5mm}
\begin{subfigure}{0.25\textwidth}
  \centering
  \includegraphics[width=\textwidth]{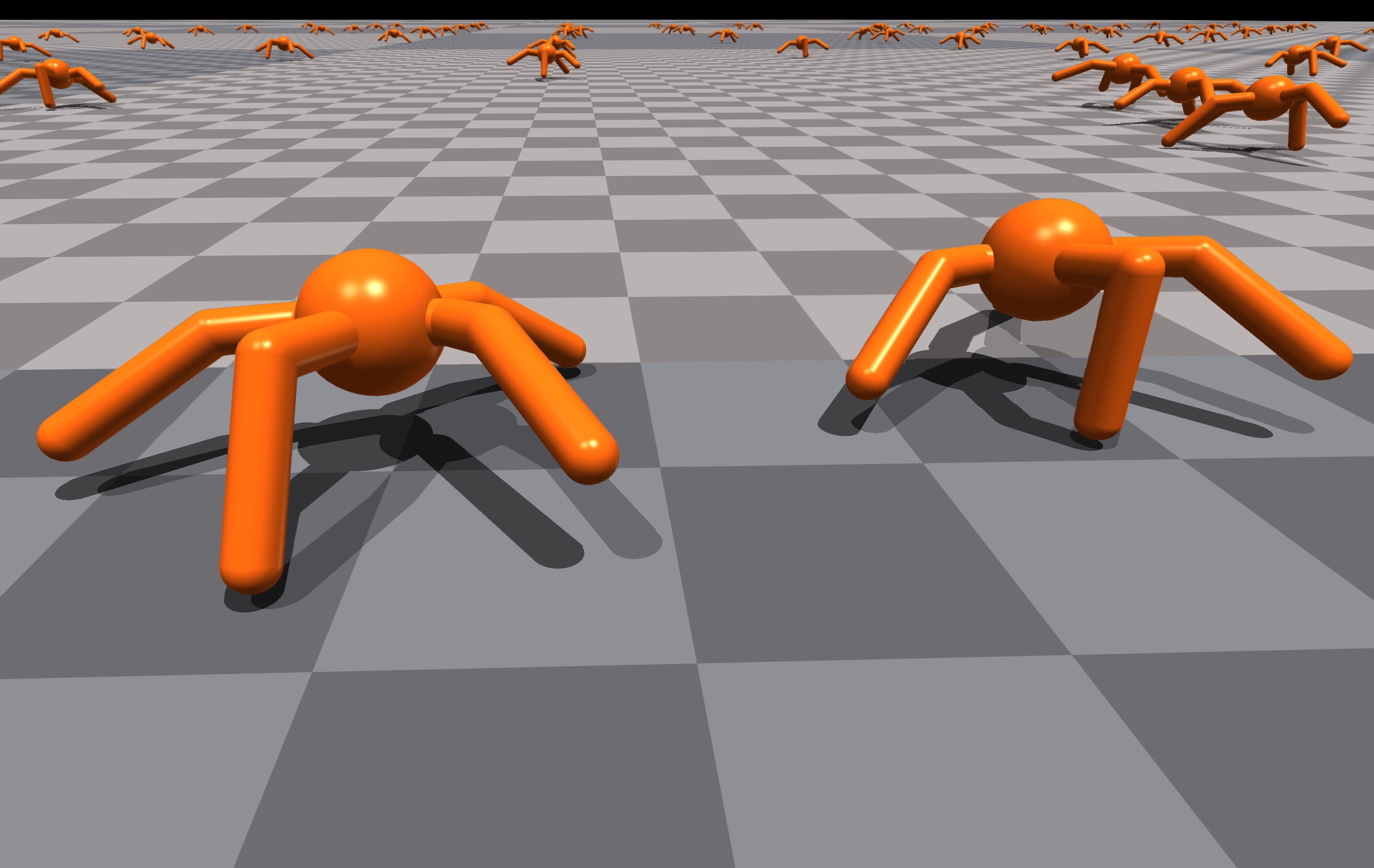} 
\end{subfigure} %
\hspace{-1.5mm}
\begin{subfigure}{.25\textwidth}
  \centering
    \includegraphics[width=\textwidth,trim={2cm, 5cm, 0cm, 0.9cm}, clip]{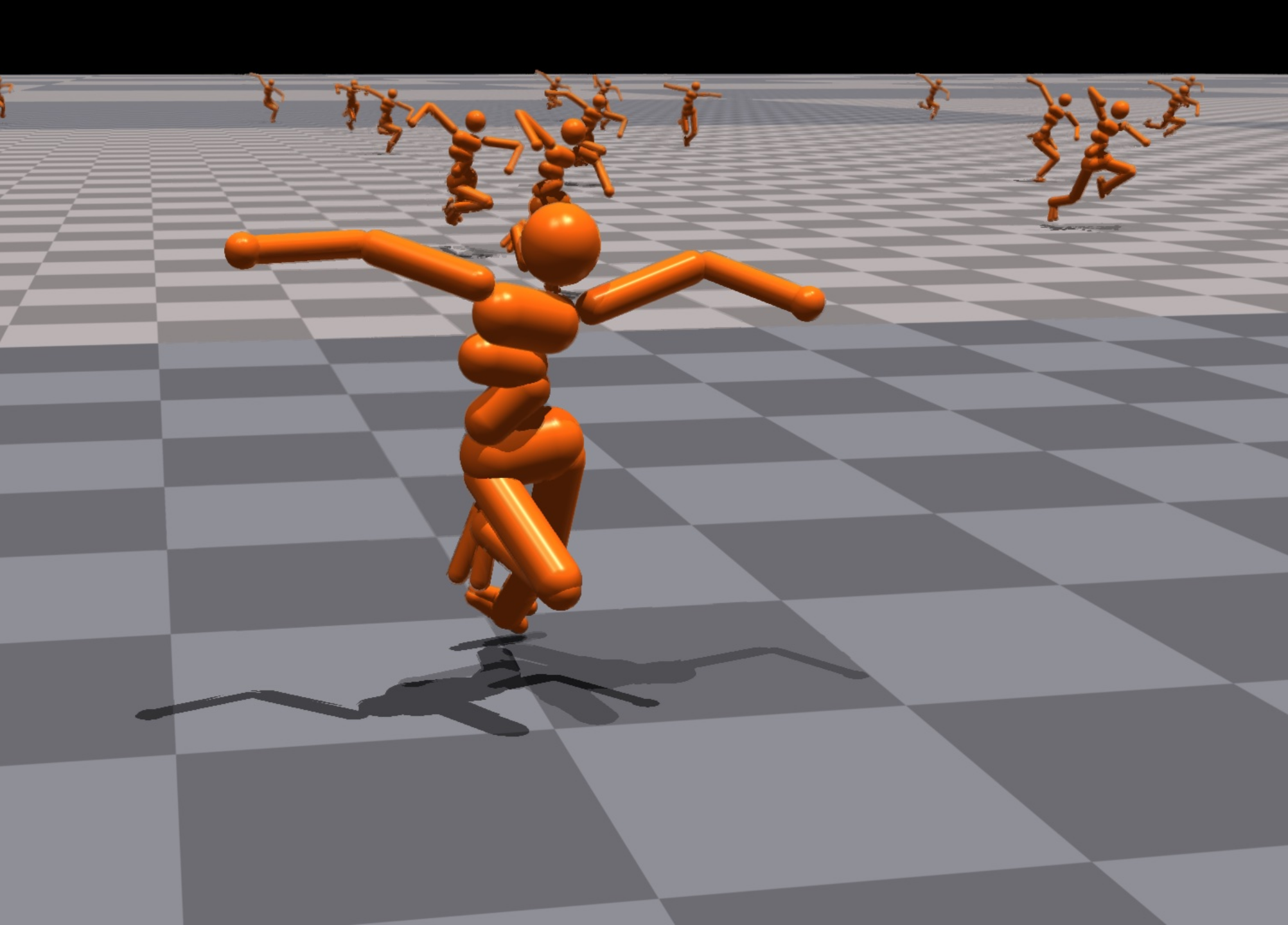}
\end{subfigure}
\hspace{-1.5mm}
\begin{subfigure}{.25\textwidth}
  \centering
  \includegraphics[width=\linewidth,trim={3cm, 3cm, 0cm, 1.0cm},clip]{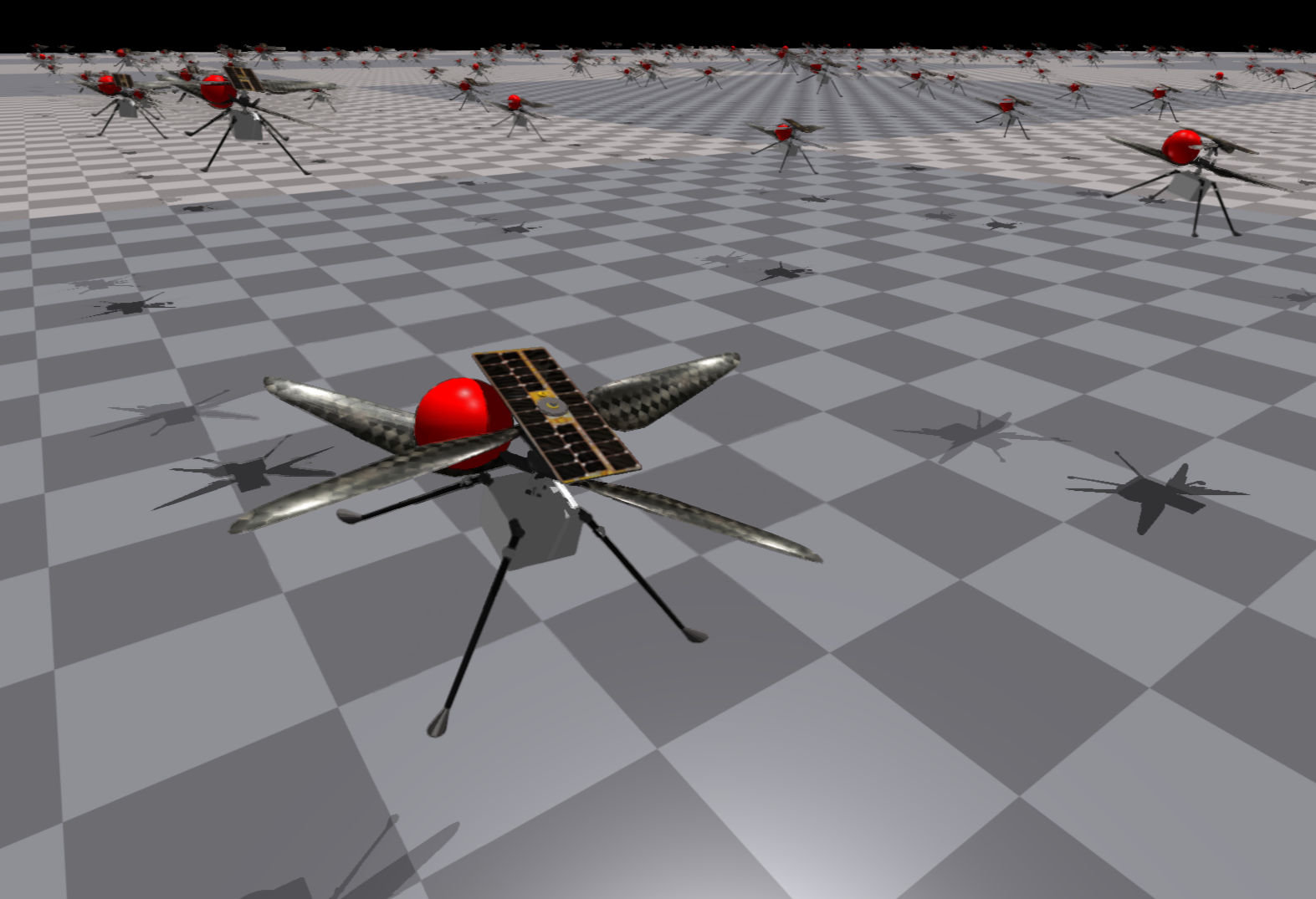}
\end{subfigure}
\hspace{-1.5mm}
\begin{subfigure}{.25\textwidth}
  \centering
    \includegraphics[width=\linewidth,trim={2.9cm, 2.9cm, 0cm, 0.2cm},clip]{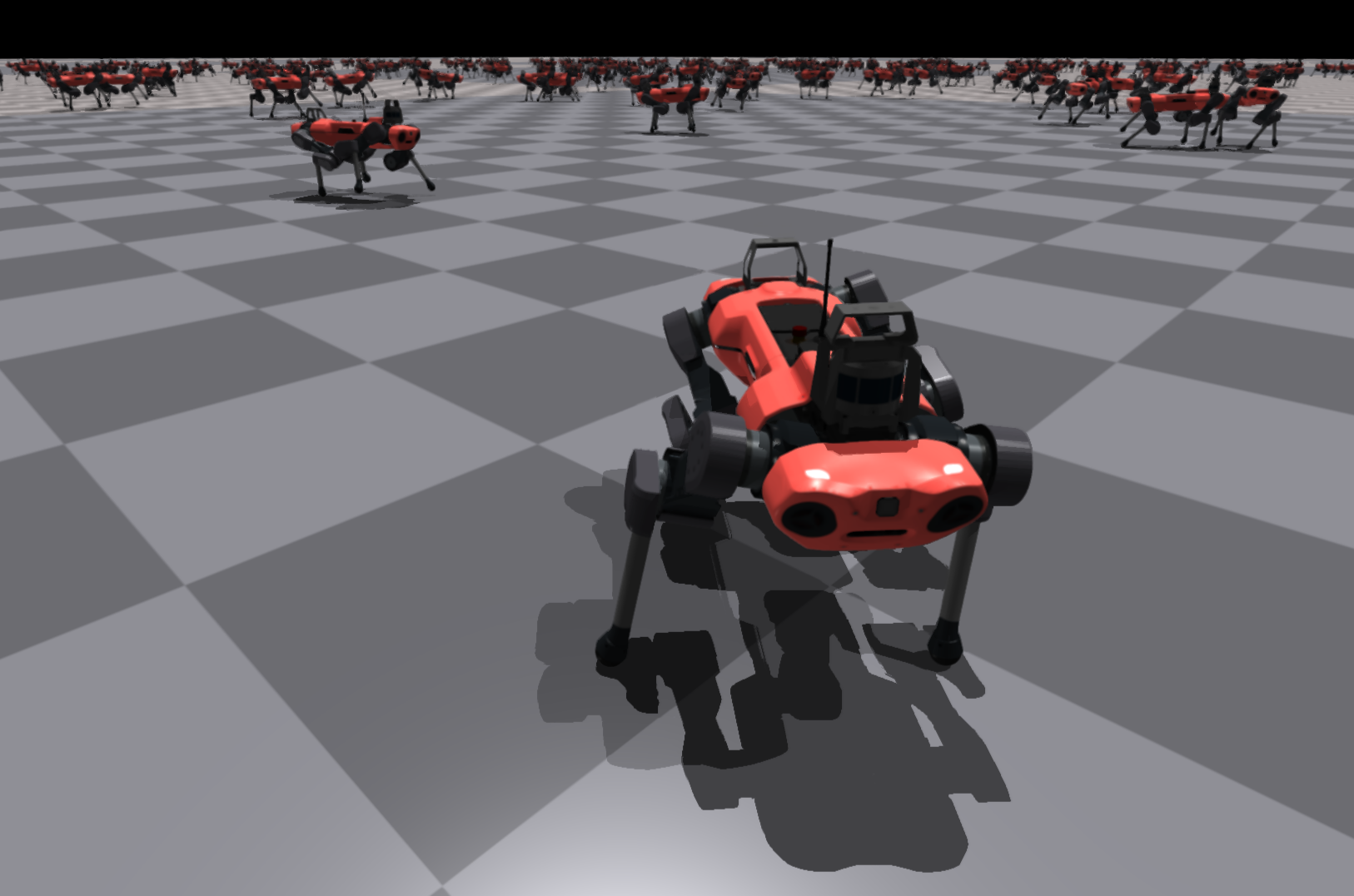}
\end{subfigure}
\newline
\hspace{-1.5mm}
\begin{subfigure}{0.25\textwidth}
  \centering
  \includegraphics[width=\textwidth]{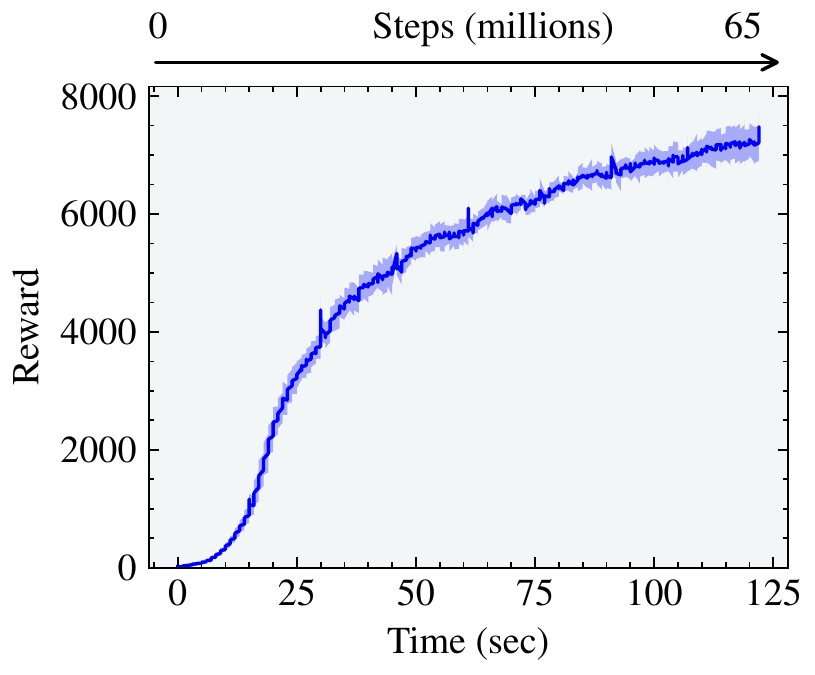}
  \caption{Ant}
\end{subfigure} %
\hspace{-1.5mm}
\begin{subfigure}{.25\textwidth}
  \centering
    \includegraphics[width=\textwidth]{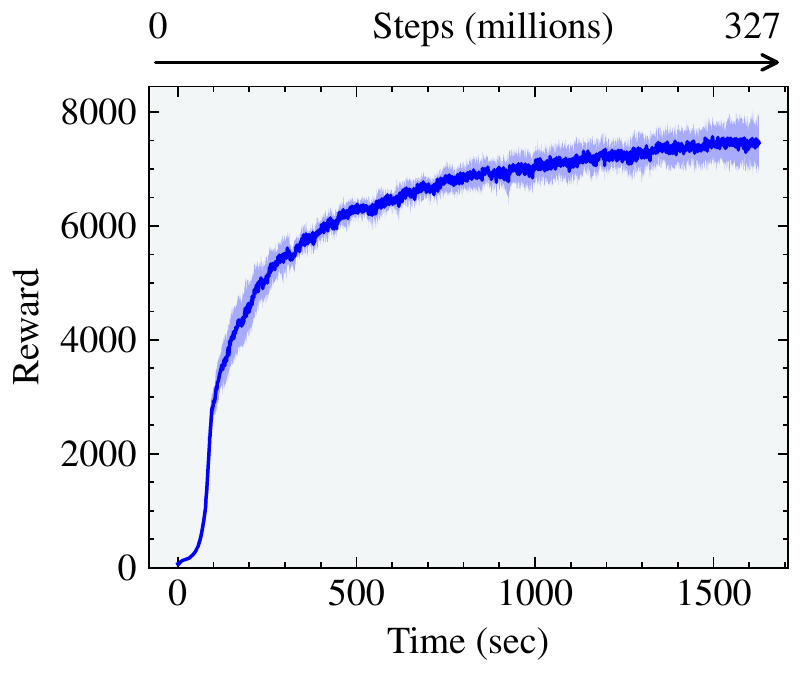}
 \caption{Humanoid}
\end{subfigure}
\hspace{-1.5mm}
\begin{subfigure}{.25\textwidth}
  \centering
  \includegraphics[width=\linewidth]{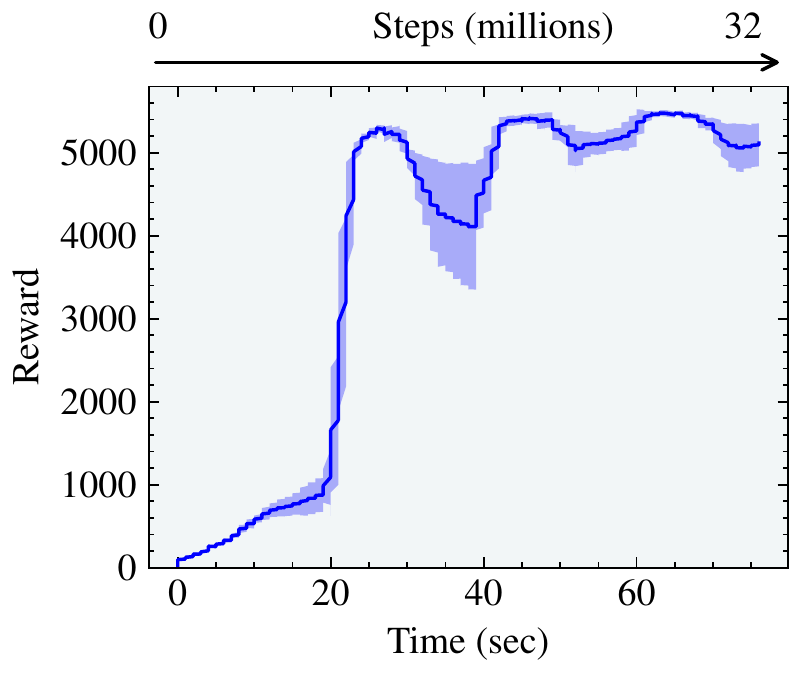}
\caption{Ingenuity}  
\end{subfigure}
\hspace{-1.9mm}
\begin{subfigure}{.25\textwidth}
  \centering
    \includegraphics[width=\linewidth]{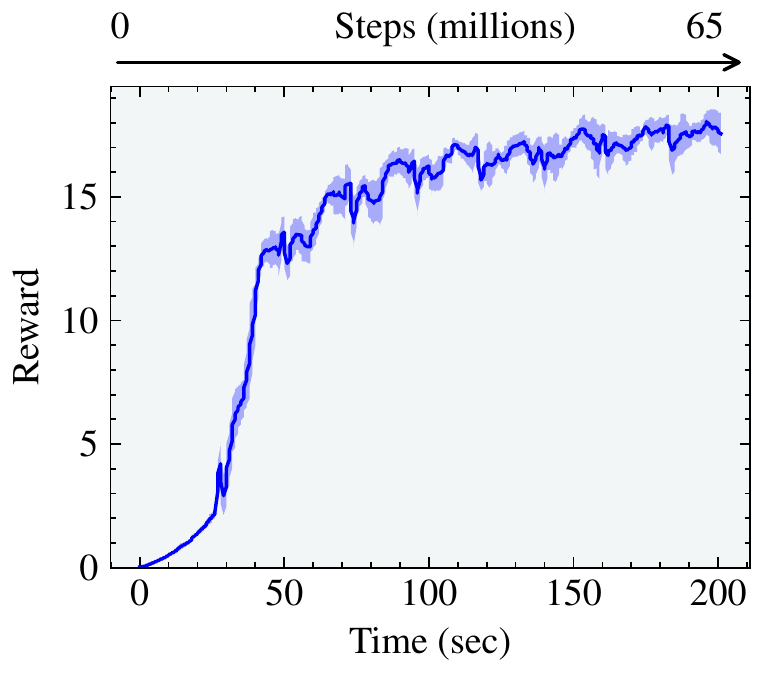}
\caption{ANYmal}      
\end{subfigure}
\caption{Locomotion environments and the corresponding reward curves.}
\label{fig:environment-ant-humanoid-ingenuity-anymal-graph}
\end{figure}

\subsection{Locomotion environments}

\subsubsection{Ant}

The Ant model has four legs with two degrees of freedom per leg. On A100 with 4096 agents simulated in parallel we find that ant can learn to run and achieve a reward above 3000 in just 20 seconds, and fully converge in under 2 minutes. The average simulation performance achieved during training is 540K environment steps per second. The results are shown in Figure \ref{fig:environment-ant-humanoid-ingenuity-anymal-graph}(a). For details of the reward function used, we refer to Appendix \ref{sec:env-details:mujoco} and for the observations used, we refer to Appendix \ref{sec:env-details:ant}.

\subsubsection{Humanoid}

The Humanoid environment has 21 DOFs and on a A100 with 4096 agents simulated in parallel we can train it to run --- a reward threshold of 5000 --- in less than 4 minutes. This is 4x faster than our previous results in \cite{Liang:etal:CoRL2018} obtained using the same threshold. As shown in Figures \ref{fig:humanoid-wrt-envs} and \ref{fig:humanoid-wrt-envs-2}, we achieve peak performance for this environment at 4096 agents. Figure \ref{fig:environment-ant-humanoid-ingenuity-anymal-graph}(b) shows the evolution of reward as a function of time. For details of the reward function used, we refer to Appendix \ref{sec:env-details:mujoco} and for the observations used, we refer to Appendix \ref{sec:env-details:humanoid}.

\subsubsection{Ingenuity}
We train a simplified model of NASA's Ingenuity helicopter to navigate to a target that periodically teleports to different locations. The environment with trained with 4096 agents and achieves a reward of 5000 in just under 30 seconds. Forces are applied directly to the two rotors on the chassis, rather than simulating aerodynamics. We use a gravity value of -3.721 $m/s^2$ to simulate martian gravity. In Figure \ref{fig:environment-ant-humanoid-ingenuity-anymal-graph}(c) we show how the reward increases as a function of time.

\subsubsection{ANYmal Robot Locomotion}

ANYmal is a robot developed by ANYbotics for industrial maintenance. It is a four-legged dog-like robot, and has been used for experiments on navigation of rough and variable terrain. We train the robot to follow target X, Y, and yaw base velocities while minimizing joint torques. The target velocities are randomized at each reset and are provided as observations alongside the positional and angular velocities of the base, the measured gravity vector, most recent actions, and DOF positions and velocities. With 4096 agents simulating in parallel, we find that the robot is able to follow the targets in under 2 minutes as shown in Figure~\ref{fig:environment-ant-humanoid-ingenuity-anymal-graph}(d). The reward function is defined in \ref{sec:env-details:anymal-locomotion}

\paragraph{ANYmal Sim-to-real on Uneven Terrain}

In addition to the simple flat terrain environment, we have developed a rough terrain locomotion task for ANYmal and validated the approach by transferring trained policies to the real robot. The robot learns to walk on uneven surfaces, slopes, stairs and obstacles. In addition to the observations of the flat terrain environment it receives terrain height measurements around the robot's base. For sim-to-real transfer we extend the reward function, add noise to the observations, randomize the friction coefficient of the ground, randomly push the robots during the episode and add an actuator network to the simulation. Following the approach used in \cite{Hwangbo_2019}, the actuator network is trained to model the complex dynamics of the series elastic actuators of the real robot.

\begin{wrapfigure}{L}{0.6\linewidth}
    \centering
    \subfloat{\includegraphics[width=0.464\linewidth, trim={0cm, 0cm, 0cm, 0cm}, clip]{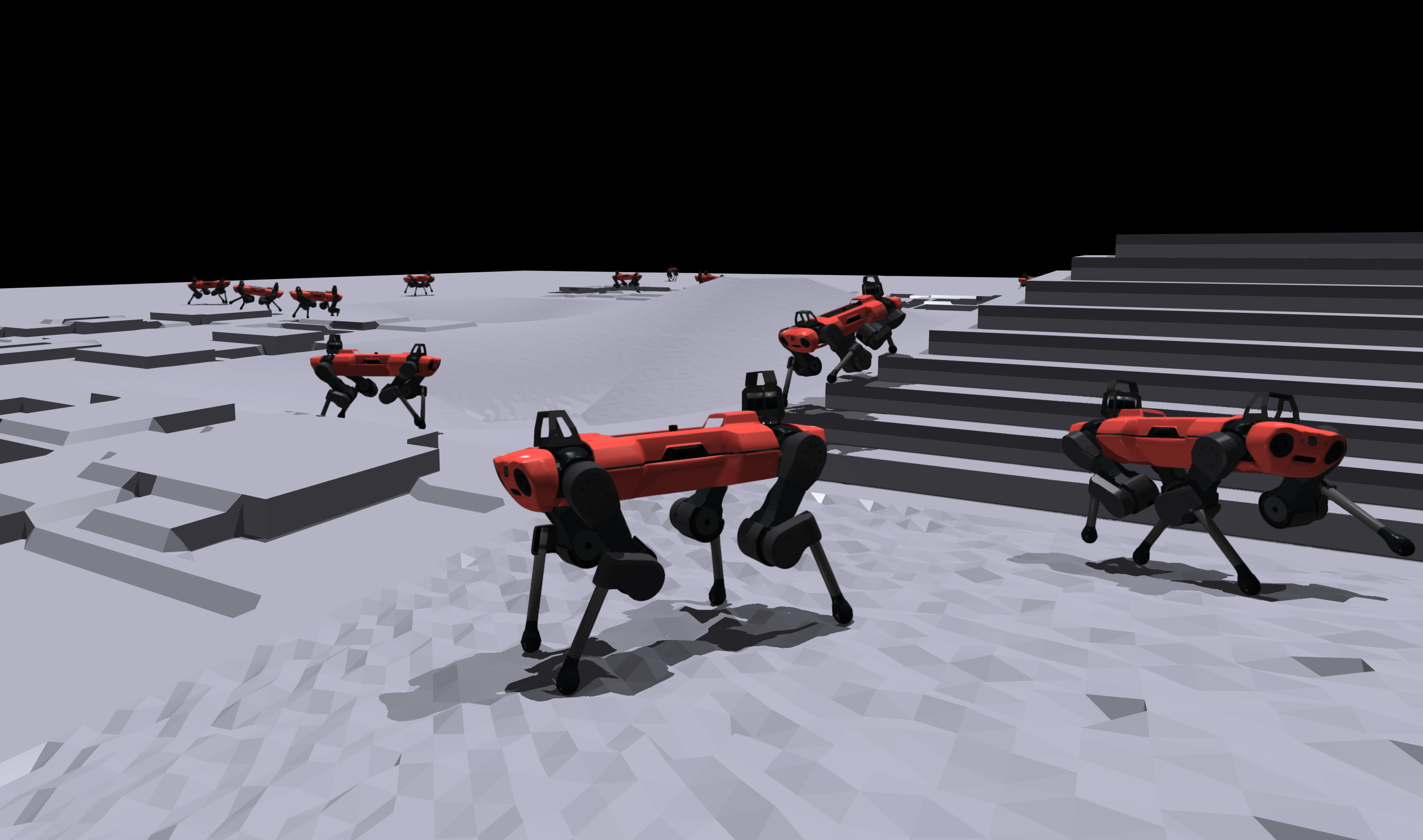}}
    \subfloat{\includegraphics[width=0.50\linewidth, trim={2cm, 7cm, 0cm, 1.5cm}, clip]{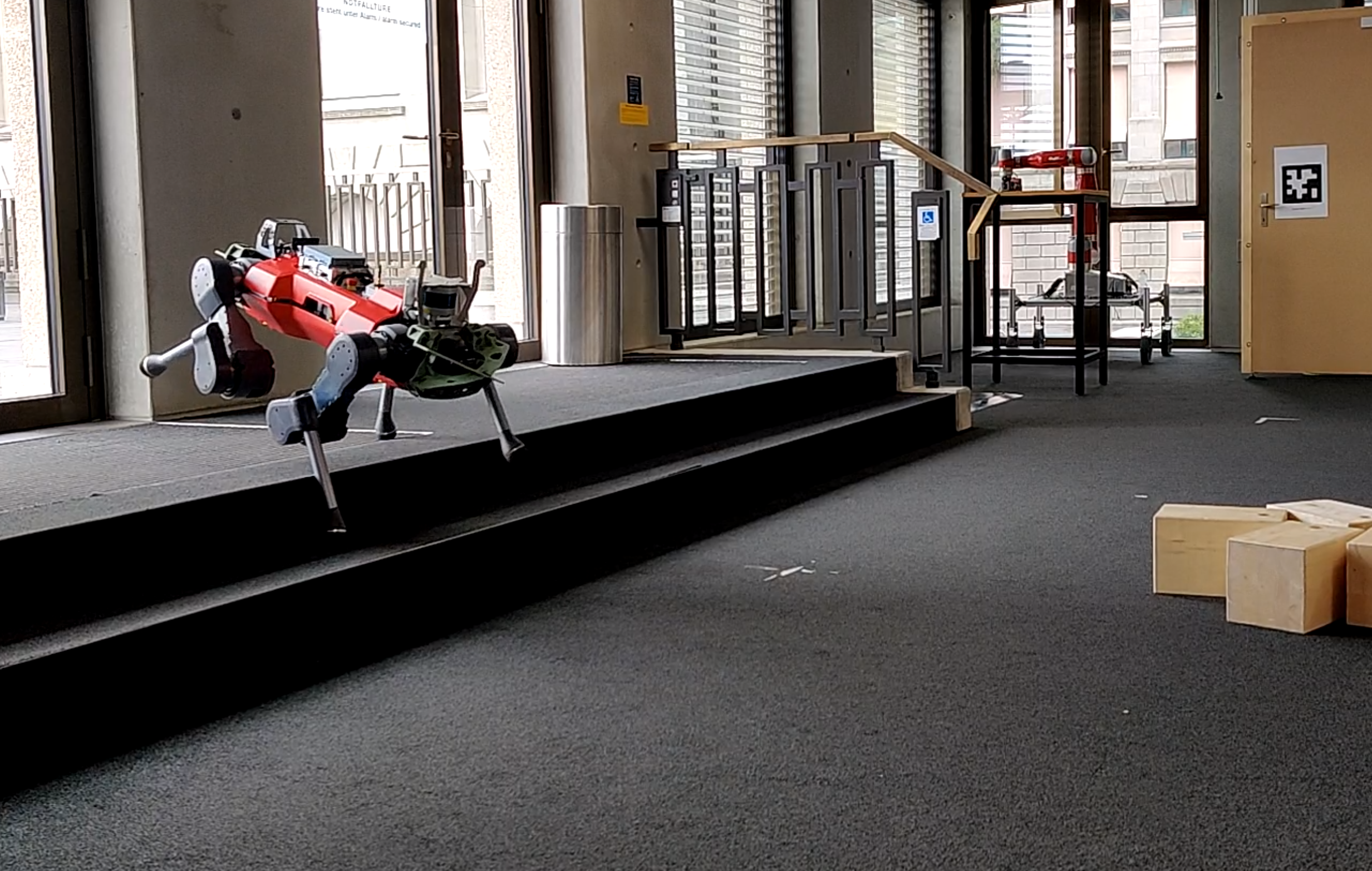}}
    \caption{Trained policy for ANYmal on rough terrain tested in simulation and on the real robot.}
    \label{fig:anymal_rough_terrain}
\end{wrapfigure}

We implement an automatic curriculum of increasing terrain difficulties. The robots start to learn on simple versions of the terrains, and when they are able to solve a certain level the difficulty is automatically increased. In order to avoid costly terrain generation during training, we create a single mesh with all terrain types and levels and change the robots' reset location depending on their progress. With 4096 environments, we can train the full task on NVIDIA RTX A6000 and transfer to the real robot in under 20 minutes. We refer to \cite{isaacgym-anymal-anonymous} for more details.

\subsection{Humanoid Character Animation}

We evaluate the performance of Isaac Gym on adversarial imitation learning tasks using an implementation of adversarial motion priors (AMP) \citep{2021-TOG-AMP}. This technique enables physically simulated humanoid character to imitate complex behaviors from reference motion data. Instead of a manually engineered imitation objective, as is commonly used in prior systems~\citep{2018-TOG-deepMimic}, AMP learns an imitation objective using an adversarial discriminator trained to differentiate between motion from the dataset and motions produced by the policy. 

\begin{figure}[h]
\centering
\includegraphics[width=1\linewidth]{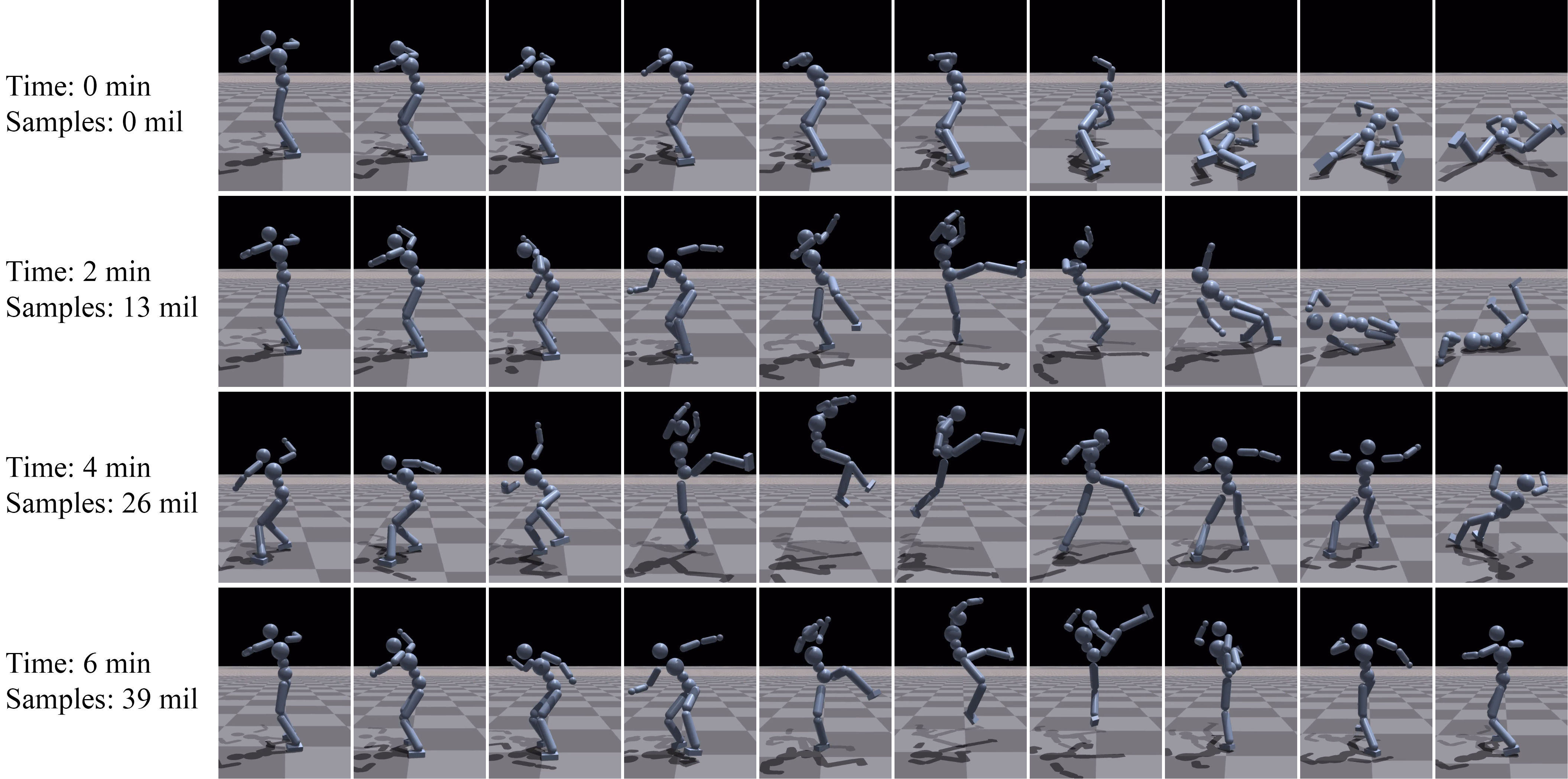}
\caption{Humanoid character trained using AMP to imitate a spin-kick.}
\label{fig:ampSpinkickTraining}
\end{figure}

Our character is modelled as a 34-DOF humanoid \cite{2021-TOG-AMP}, and all motion clips are recorded from human actors using motion capture. Table~\ref{tab:env-details:amp:obs} in Appendix \ref{sec:env-details:amp} details the observation features. 
The adversarial training process enables the character to closely imitate a diverse corpus of motions, ranging from common locomotion behaviors, such as walking and running, to more athletic behaviors, such as spin-kicks and dancing. Effective policies can be learned with approximately 39 million samples, requiring approximately \textbf{6 minutes} 
with 4096 environments. The implementation provided by Peng \textit{et al.}, 2021 \cite{2021-TOG-AMP} requires about \textbf{1 day} (30 hours) on 16 CPU cores to simulate a similar number of samples in PyBullet. Therefore, Isaac Gym provides \textbf{300x or 2.48 orders of magnitude improvement} in the training time.

\subsection{Franka Cube Stacking}

\begin{wrapfigure}{L}{0.70\textwidth}
\vspace{-7mm}
\centering
\hspace{-7mm}
\begin{subfigure}{.41\linewidth}
  \vspace{-6mm}
  \centering
  \includegraphics[width=0.85\linewidth]{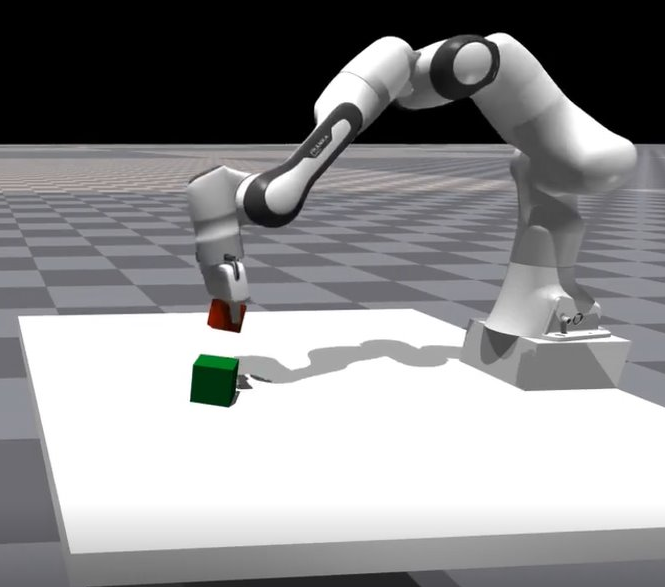}
  \label{fig:franka_cube_stack-pic}
\end{subfigure} %
\hspace{-5mm}
\begin{subfigure}{.5\linewidth}
  \centering
  \includegraphics[width=\linewidth]{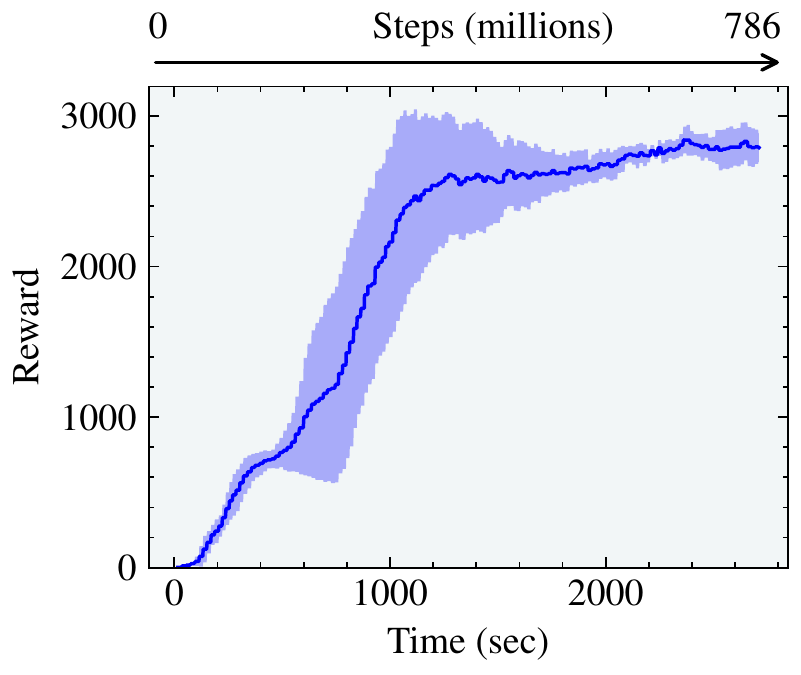}
  \label{fig:franka_cube_stack-training}
\end{subfigure}
\vspace{-5mm}
\caption{The Franka Cube Stacking environment and the corresponding reward curves.}
\vspace{-2mm}
\label{fig:franka_cube_stack}
\end{wrapfigure}

We use 16384 agents to train a Franka robot to stack a cube on top of an other. In this environment, we use a slightly different choice of action space, Operation Space Control (OSC), for learning. OSC ~\cite{khatib1987osc} is a task-space compliant controller that has been shown to enable faster policy learning compared to joint-space controllers ~\cite{zhu2020robosuite} and learn contact-rich tasks ~\cite{martin-martin2019vices}. Our OSC implementation is fully differentiable in Isaac Gym and we obtain convergence with this controller in under 25 minutes. Figure~\ref{fig:franka_cube_stack} shows the training results.

\subsection{Robotic Hands}

\begin{figure}[h]
\centering
\begin{subfigure}[c]{0.333\textwidth}
    \centering
    \includegraphics[width=0.99\textwidth]{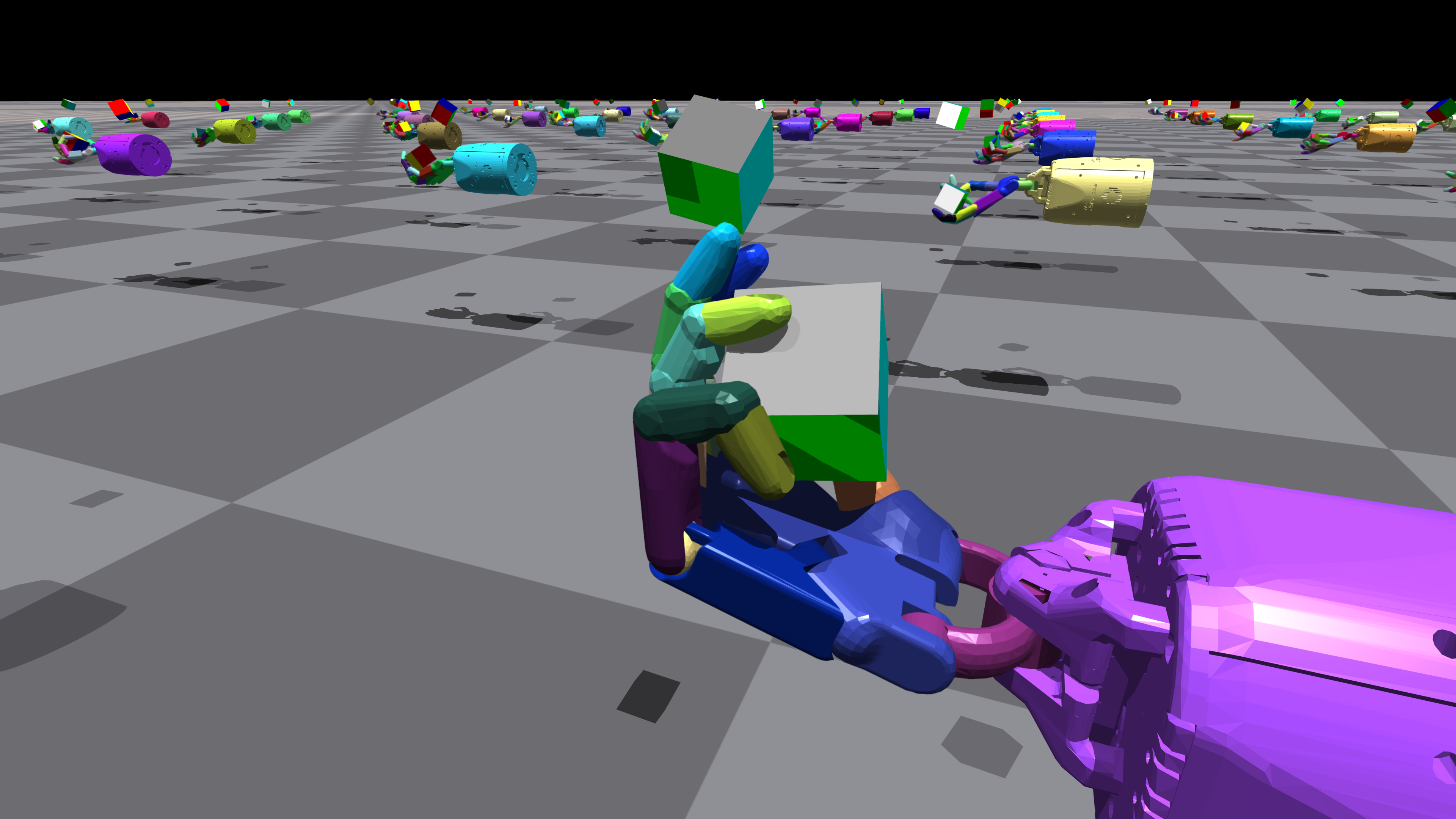}
\end{subfigure}%
\begin{subfigure}[c]{0.333\textwidth}
    \centering
    \includegraphics[width=0.99\textwidth]{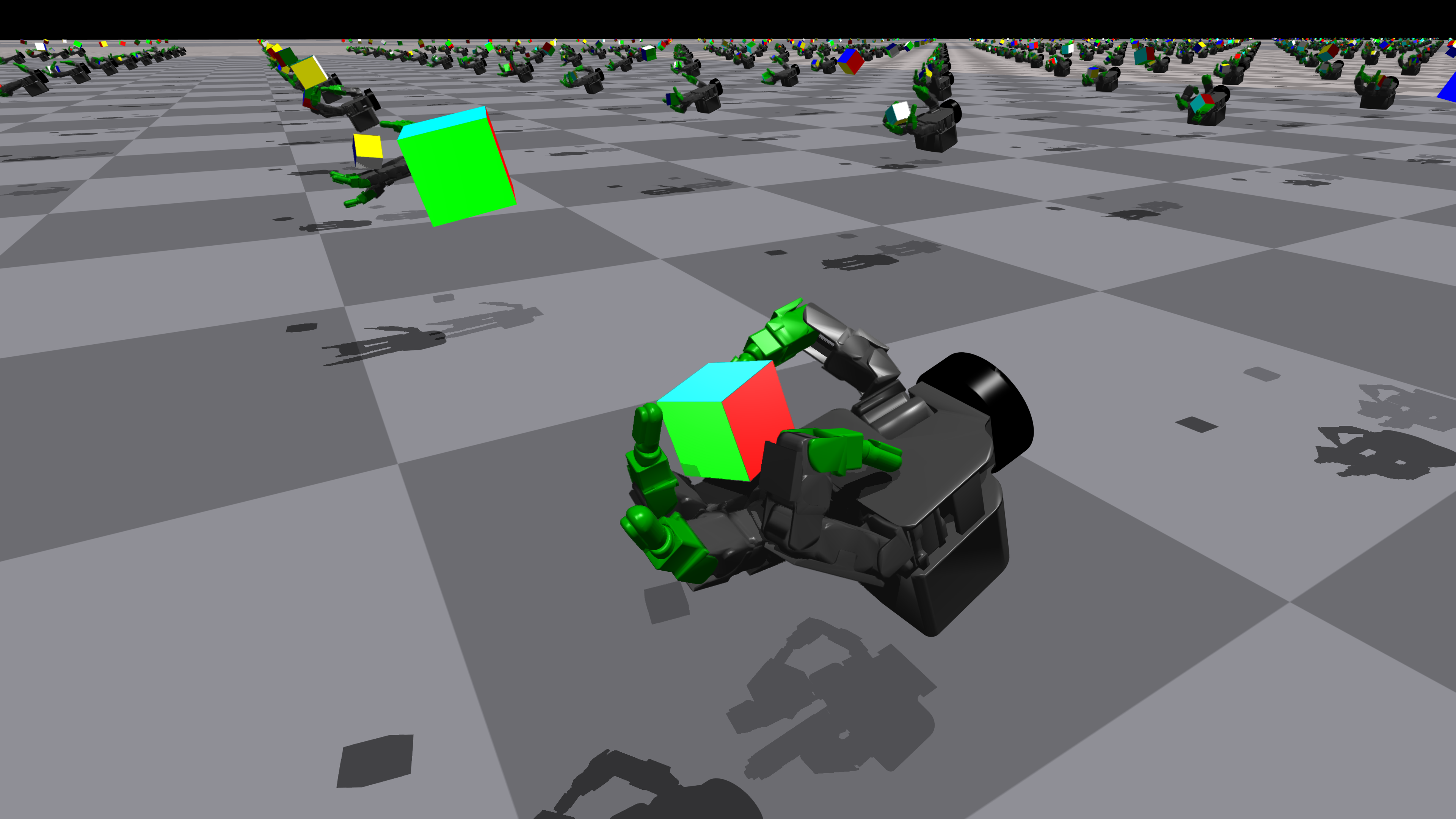}
\end{subfigure}%
\begin{subfigure}[c]{0.333\textwidth}
    \centering
    \includegraphics[width=0.99\textwidth]{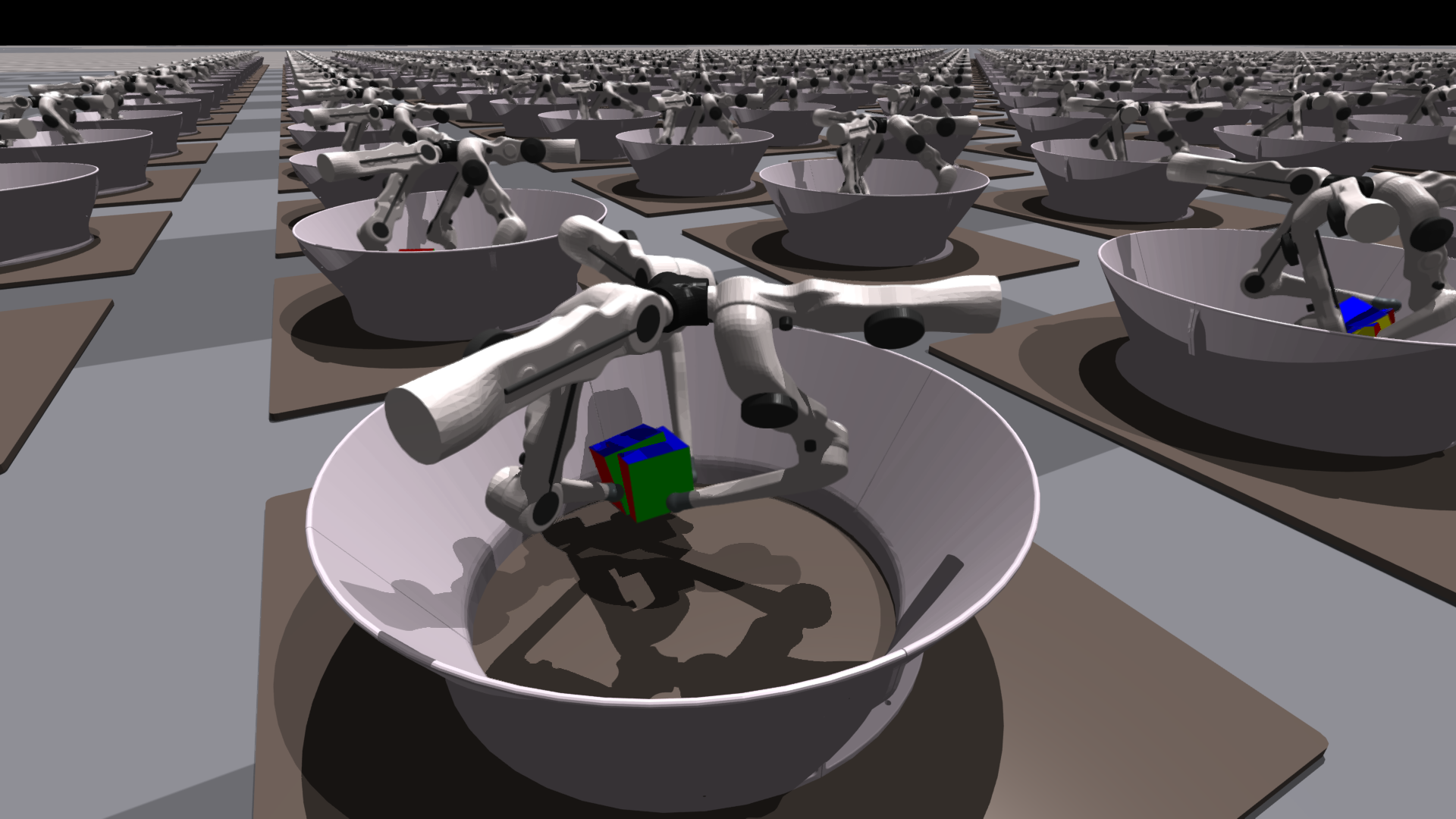}
\end{subfigure}%
\caption{The three in-hand manipulation environments implemented in Isaac Gym: Shadow Hand, Trifinger, and Allegro.}
\label{fig:hands}
\end{figure}

\begin{figure}[t]
\centering
\begin{subfigure}[l]{0.25\textwidth}
    \centering
    \includegraphics[width=0.98\textwidth]{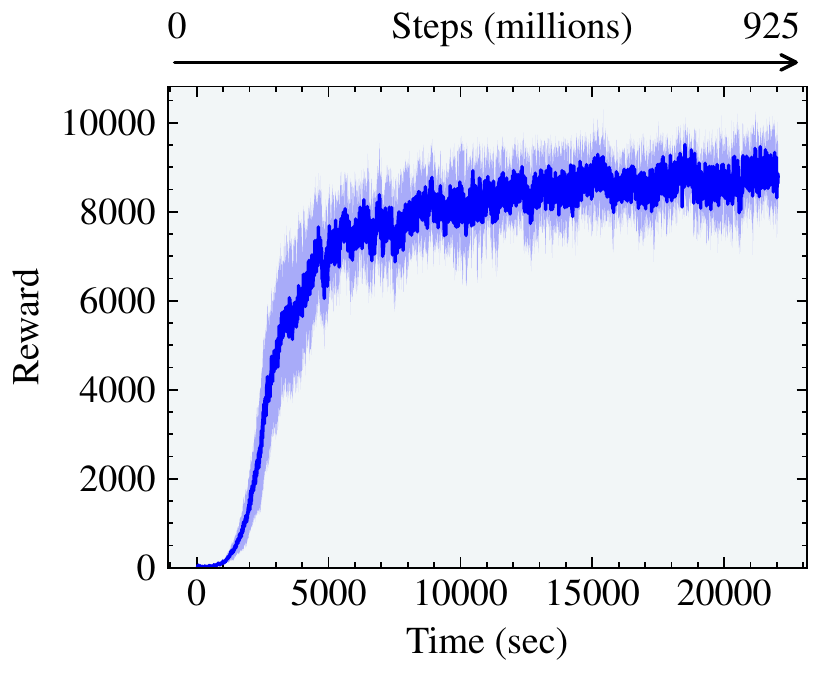}
    \caption{SH OpenAI LSTM}
\end{subfigure}%
\begin{subfigure}[c]{0.25\textwidth}
    \centering
    \includegraphics[width=0.98\linewidth]{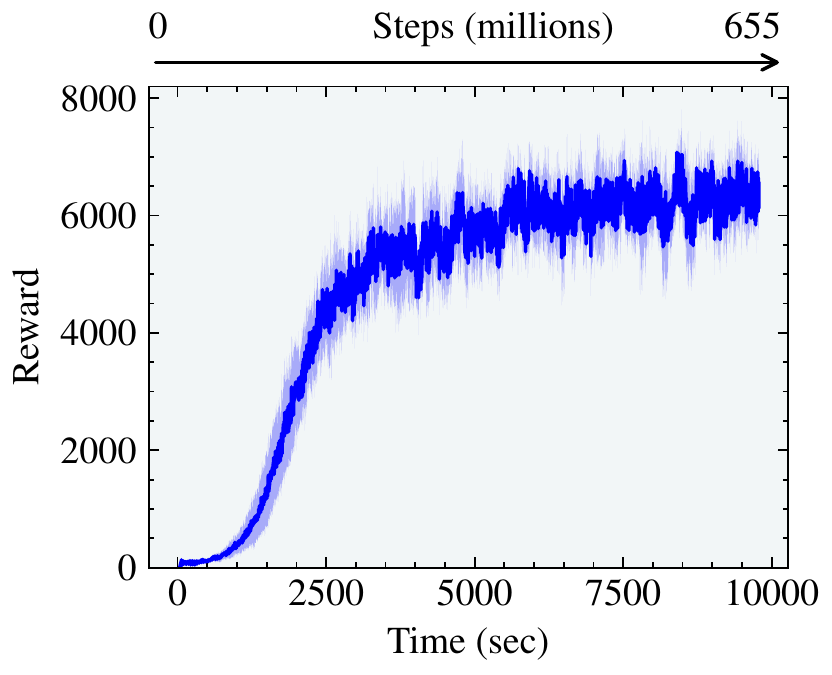}
    \caption{SH OpenAI FF}
\end{subfigure}%
\centering
\begin{subfigure}[c]{0.25\textwidth}
    \centering
    \includegraphics[width=0.94\textwidth]{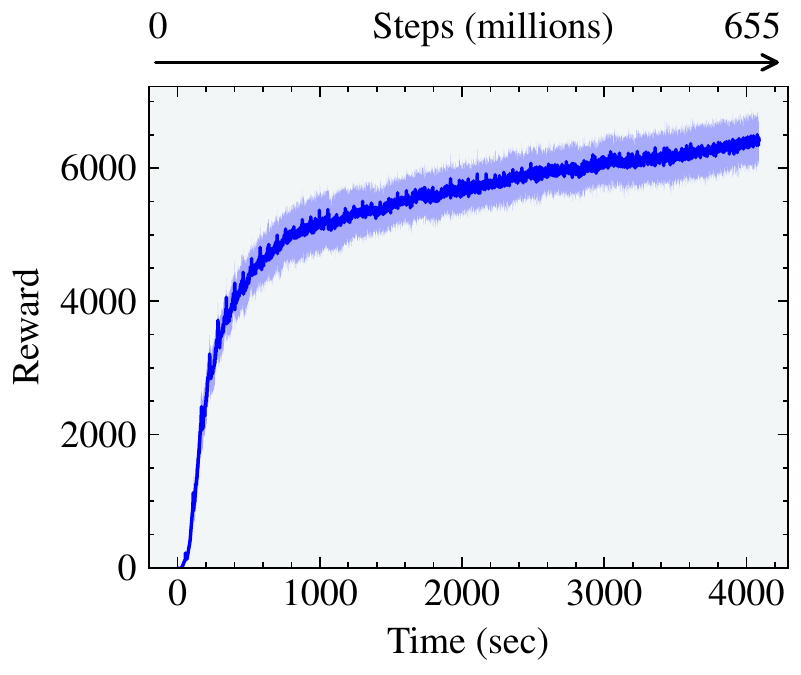}
    \caption{SH Standard}
\end{subfigure}%
\begin{subfigure}[c]{0.25\textwidth}
    \centering
    \includegraphics[width=0.94\textwidth]{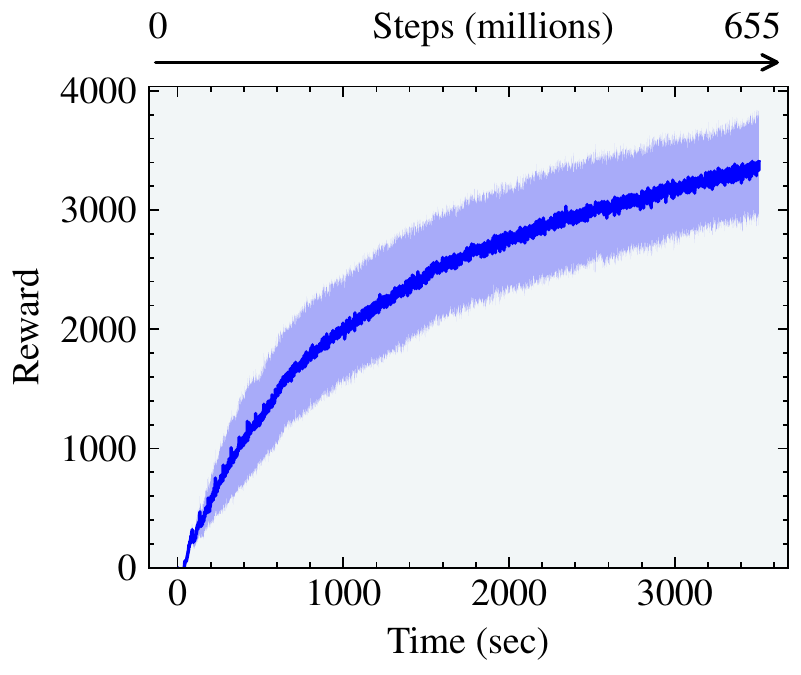}
    \caption{Allegro Hand Standard}
\end{subfigure}%
\caption{Reward curves for the three in-hand manipulation environments implemented in Isaac Gym. These results are obtained with \textbf{(a)} Shadow Hand with OpenAI observation and LSTMs, \textbf{(b)} Shadow Hand with OpenAI observation and feed forward networks \textbf{(c)} Shadow Hand with Standard observations and \textbf{(d)} Allegro Hand with Standard observations. Shadow Hand OpenAI is trained with asymmetric actor-critic and domain radomisation while Shadow Hand Standard and Allegro Hand Standard are trained with standard observations and symmetric actor-critic with no domain randomisation.}
\vspace{-3mm}
\label{fig:hands-rewards}
\end{figure}

Large-scale simulation has the ability to solve not just individual instances but whole classes of problems in robotics, by leveraging the generality of the model-free reinforcement learning framework. Dexterous manipulations is one of the most challenging problems in robotics. 

To show the performance of our simulator and the ability to realistically model contact we implemented 3 different hand training environments as shown in Figure\ref{fig:hands}. Shadow Hand and Allegro Hand are trained to learn cube orientation while TriFinger learns to repose the cube in 6 degrees-of-freedom involving rotation and translation. We now focus on the specific training details for these environments.

Firstly, the Shadow Dexterous Hand. We follow the standard formulation where policy and value function both receive the same input as well as OpenAI observations with asymmetric formulation and domain randomisation from \citep{openai-sh}. Secondly, the TriFinger robot \citep{trifinger-platform}, which shows the ability to do 6-DoF manipulation by reposing the cube to a desired position and orientation, a task which has previously shown to be challenging for model-free reinforcement learning \citep{isaacgym-trifinger}. We use asymmetric actor-critic and domain randomisation for TriFinger and demonstrate sim-to-real transfer on a real robot. Finally, we reuse system from the Shadow Hand to the Allegro hand with minimal changes to show the generality of our approach. These three environments are depicted in Figure \ref{fig:hands} and the corresponding reward curves in Figure \ref{fig:hands-rewards}.

\newpage

\subsubsection{Shadow Hand}
\label{sec:env:shadow-hand}
As mentioned, the task with Shadow Hand is to manipulate the cube to achieve a specific target orientation and is inspired by OpenAI \textit{et al.} \cite{openai-sh}. We train with multiple variants on the Shadow Hand environment and describe them below:

\paragraph{Shadow Hand Standard}

In this setting, we use a standard formulation for training where the policy and the value function use feed forward networks and  receive the same input observations. 
The default observations we used for the Shadow Hand Standard include joint position, velocities, forces, force-torque sensors reading from each fingertip, manipulated object position and orientation, linear and angular velocities, goal orientation, relative rotation between the current object and target rotations, actions applied on the previous step. For a detailed overview of observation and reward, see Appendix ~\ref{sec:shadow-hand-details}. Also note that this variant does not use any randomisations.

\paragraph{Shadow Hand OpenAI}

We also reproduce results with OpenAI Shadow Hand experiments in Isaac Gym with observations used in dexterity work from OpenAI \textit{et al.} \cite{openai-sh}. A key difference between this and the Shadow Hand Standard variant is that it uses asymmetric observations. The policy receives only the input observations that are possible to obtain in the real world settings while the value function receives the same observations in addition to the other privileged information available from the simulator. This variant should make it possible to transfer the policy to the real world, mimicking the setup in \cite{openai-sh}. The observations for the policy and value function are provided in Table~\ref{tab:env-details:shadow-openai:obs}. We experiment with both feed forward networks (SH OpenAI FF) and LSTMs (SH OpenAI LSTM). The LSTM networks are trained with a sequence length of 4. 

It is worth noting that only networks trained with OpenAI observations use domain randomisation to closely match the results in OpenAI dexterity work \cite{openai-sh}.

\begin{figure}[h]
\centering
\begin{subfigure}[c]{0.25\textwidth}
    \centering
    \includegraphics[width=0.98\textwidth]{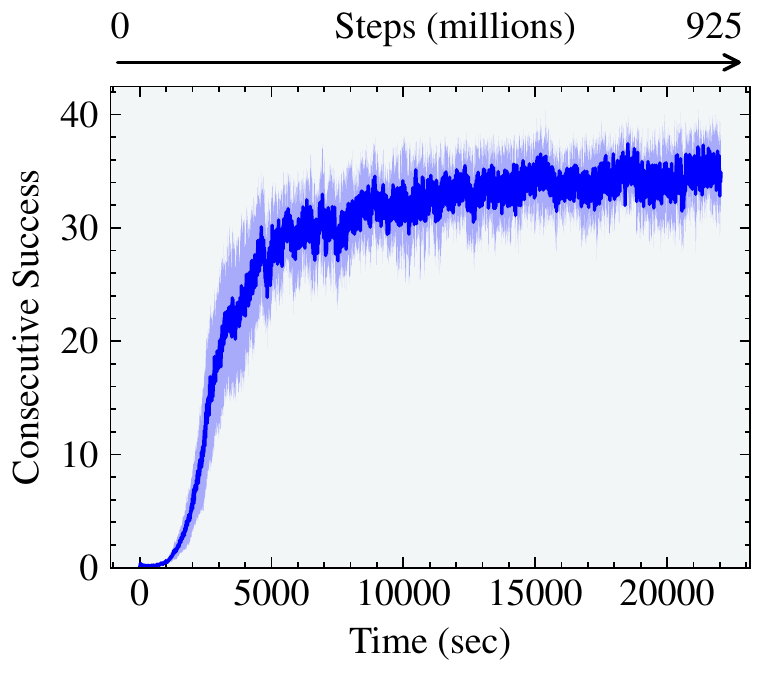}
    \caption{SH OpenAI LSTM}
\end{subfigure}%
\begin{subfigure}[c]{0.25\textwidth}
    \centering
    \includegraphics[width=0.98\textwidth]{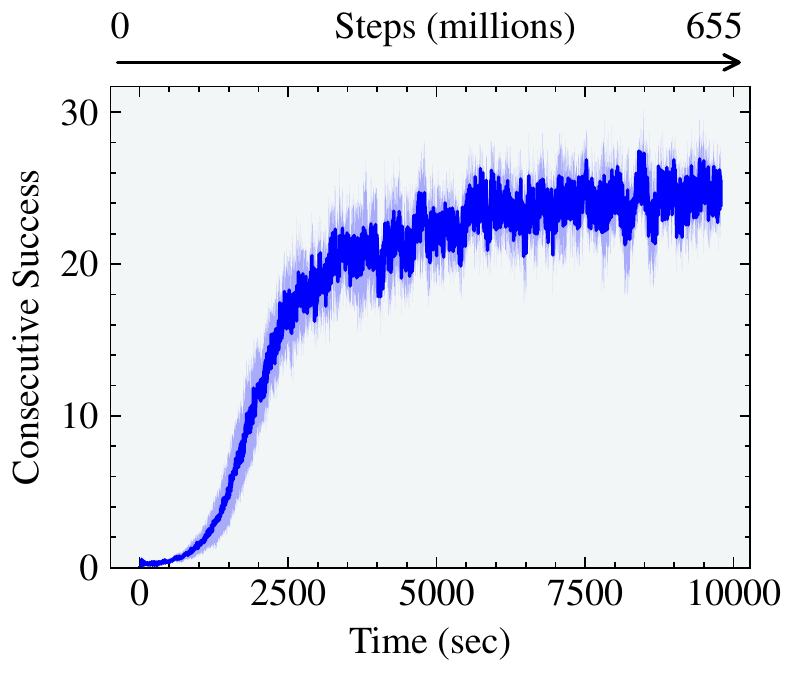}
    \caption{SH OpenAI FF}
\end{subfigure}%
\begin{subfigure}[c]{0.25\textwidth}
    \centering
    \includegraphics[width=0.95\textwidth]{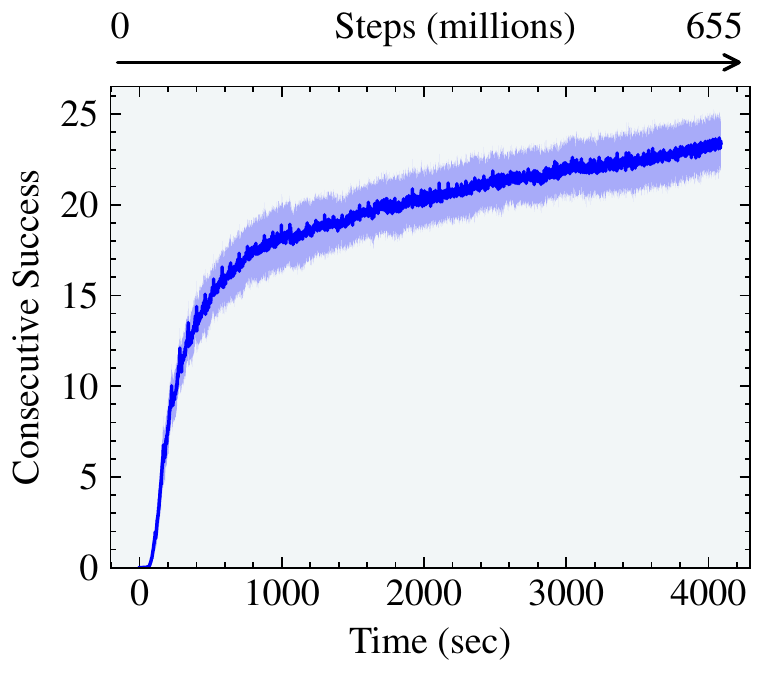}
    \caption{SH Standard}
\end{subfigure}%
\begin{subfigure}[c]{0.25\textwidth}
    \centering
    \includegraphics[width=0.97\textwidth]{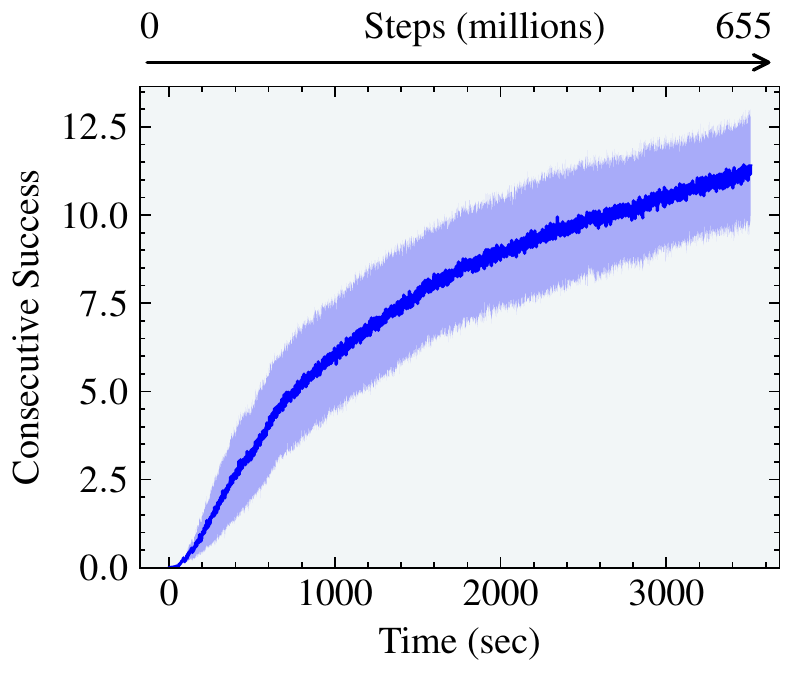}
    \caption{Allegro Hand Standard}
\end{subfigure}%
\caption{Consecutive successes per episode for \textbf{(a)} Shadow Hand with OpenAI observation and LSTMs, \textbf{(b)} Shadow Hand with OpenAI observation and feed forward networks \textbf{(c)} Shadow Hand with Standard observations and \textbf{(d)} Allegro Hand with Standard observations. Shadow Hand Standard and Allegro Hand Standard both use feed forward networks for policy and value functions.}
\label{fig:hands:consecutive-success}
\end{figure}

\paragraph{Randomizations}
For domain randomization we closely followed the approach proposed in \citep{openai-sh} and applied correlated and uncorrelated noise to observations, actions, as well as randomized cube size and all the key physics properties -- masses, inertia tensors, friction, restitution, joint limits, stiffness and damping. Full details of these are available in Appendix \ref{sec:shadow-appendix:randomizations}. 

We outline a few important differences between our setup and the one used in the OpenAI work below: 
\begin{itemize}
   \item While OpenAI used a success tolerance of 0.4 rad\footnote{page 22, section C.1, paragraph \textbf{Goals} in \cite{openai-sh}} \cite{openai-sh}, we use both 0.4 rad and a tighter tolerance of 0.1 rad. We focus on results with 0.4 rad in this section and provide results with 0.1 tolerance in Appendix \ref{section-sh-openai-01-tol}
   \item We use a continuous as opposed to a discrete control space used in \cite{openai-sh}.  
   \item Our results are averaged with 5 seeds while OpenAI show results with only 1 seed\footnote{page 11, section 6.3, \textbf{Ablation of Randomizations}, Figure 8 in \cite{openai-sh}}. 
   \item The randomizations used in our work do not include action delay and motor backlash.
   \item We use an LSTM layer of 1024 hidden units after the input followed by an MLP layer of 512 hidden units. On the other hand OpenAI \textit{et al.} \cite{openai-sh} used an MLP layer of size 1024 after the input followed by an LSTM layer of size 512 hidden units. We found our setting performs better with Isaac Gym.
  \item We use a somewhat different reward function to OpenAI as shown in Appendix \ref{sec:robotic-hands-rewards}.
   \item Our experiments are only in simulation and unlike \cite{openai-sh} we do not attempt any sim-to-real transfer for the Shadow Hand experiment.

\end{itemize}

Figure~\ref{fig:hands-rewards}(a), (b) and (c) show the reward curves for various settings we used for Shadow Hand. Shadow Hand Standard --- trained with no randomization and uses symmetric actor critic setting with a feed forward network --- is the fastest to reach a reward of 6000. This setting achieves 20 consecutive successes in under 35 minutes. Important to remember that this setting is not suitable for sim-to-real transfer as it includes some observations that may not be directly available in the real world. 

We now focus on experiments with OpenAI observations and asymmetric feed-forward actor-critic. This setting is suited for sim-to-real transfer and the policy uses only the observations that are possible to obtain in the real world. As shown in Figure~\ref{fig:hands:consecutive-success}(b), we achieved more than 20 consecutive successes in less than 1 hour. In contrast, for the same performance it takes 30 hours on the OpenAI setup consisting of CPU based simulation and training setup running MuJoCo \cite{MuJoCo:etal:2012} simulator on a cluster of 384 16-core CPUs with 6144 CPU cores in total and using 8 NVIDIA V100 GPUs for training.  In Figure~\ref{fig:hands:consecutive-success}(a) we show that using LSTM networks, the performance increases and we can reach 37 consecutive successes in just under than 6 hours while OpenAI \textit{et al.} \cite{openai-sh} achieve same performance in \textasciitilde{17} hours. Since OpenAI \textit{et al.} \cite{openai-sh} show results only with 1 seed, comparing their result with our best seed we note that 37 consecutive successes with LSTM experiments can be achieved in just 2.5 hours.  We provide the results for Shadow Hand OpenAI experiment with success tolerance of 0.1 in the Appendix \ref{sec:shadow-hand-details}.

\begin{wrapfigure}{L}{0.5\textwidth}
\vspace{-5mm}
    \centering
    \begin{subfigure}[c]{0.5\linewidth}
    \centering
 \includegraphics[width=\linewidth]{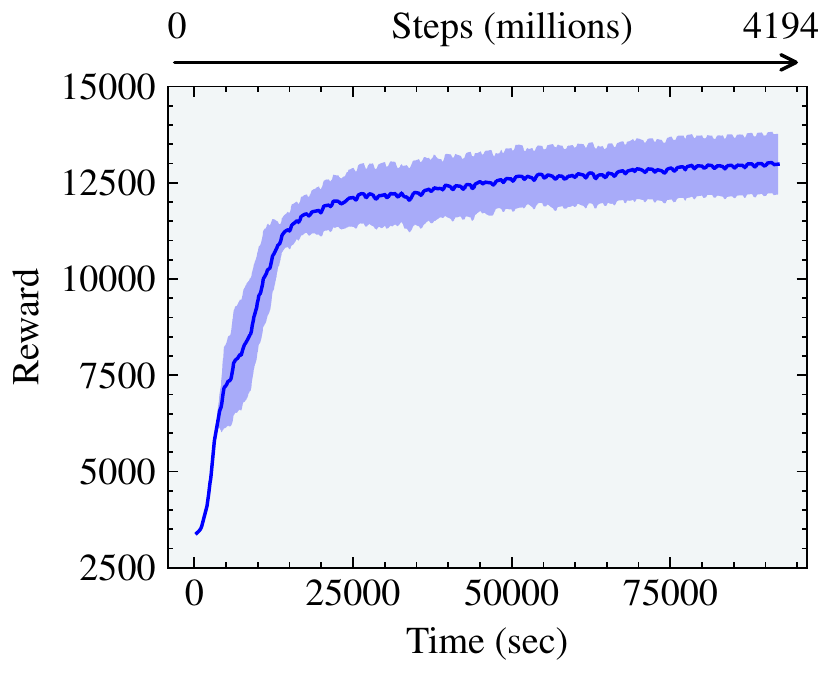}
    \caption{Reward}
\end{subfigure}%
 \begin{subfigure}[c]{0.5\linewidth}
    \centering
    \includegraphics[width=0.95\linewidth]{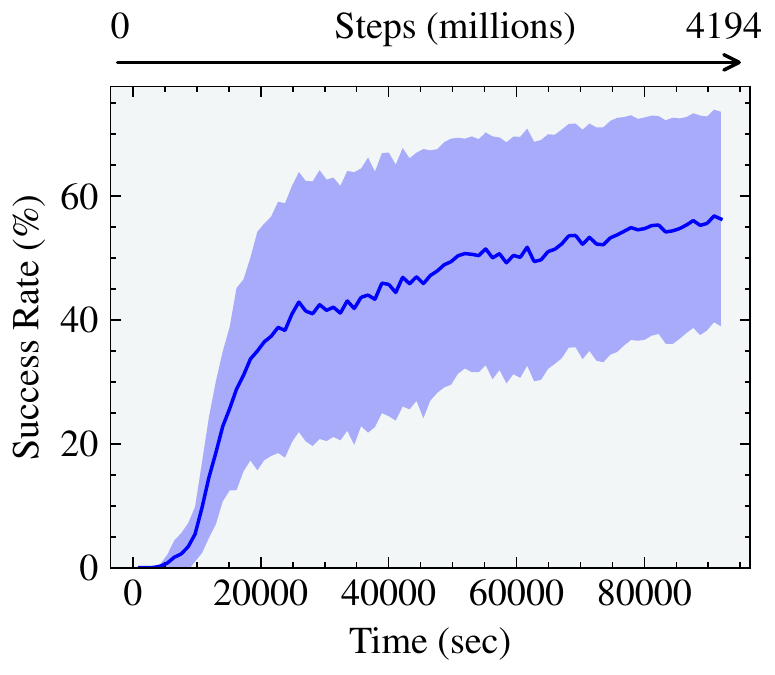}
    \caption{Success Rate}
\end{subfigure}%
    \caption{TriFinger reward and the corresponding success rate.}
\vspace{-3mm}    
\label{trifinger-reward-sr}
\end{wrapfigure}

\subsubsection{TriFinger}

The TriFinger manipulation task, originating in \citep{trifinger-platform}, involves picking a cube lying on a flat surface and repositioning it to a desired 6-degrees-of-freedom pose. The manipulator has 3 fingers each with three degrees of freedom. In \citep{isaacgym-trifinger}, it was shown that Isaac Gym training combined with Domain Randomization allows sim-to-real transfer. The environment is shown in Figure \ref{fig:hands}. 

We use an asymmetric actor-critic formulation for this system as that allows to design a policy that uses input observations that are possible to obtain in the real world and therefore enable sim-to-real transfer. We show the reward and success rate in simulation in Figure~\ref{trifinger-reward-sr}. We also transfer results from simulation to the real world and note that our mean success rate in the real world is 55\%. We refer to \cite{isaacgym-trifinger} for more detailed analysis. 

\begin{figure}[t]
\centering
\includegraphics[width=1\linewidth]{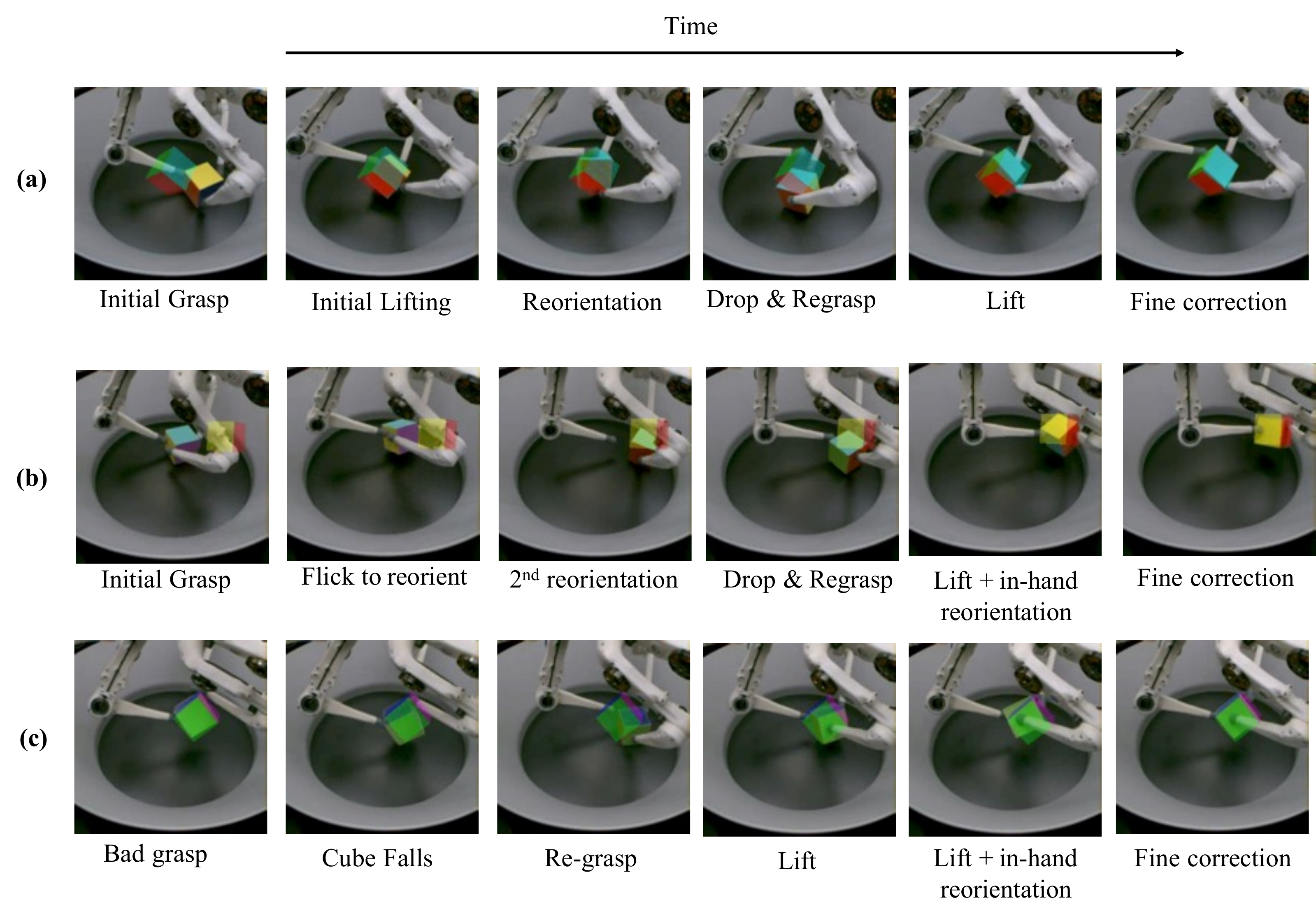}
\caption{Trifinger learns a variety of dexterous manipulation behaviours in order to move the cube to the correct position and orientation. These results are obtained on the real TriFinger robot hosted by \cite{trifinger-benchmarking, trifinger-platform}.}
\label{fig:trifingerRealRobot}
\end{figure}

In particular, this example shows the ability of policies learned using Isaac Gym's physics to generalize to the real world. Some of the behaviours leaned by the policy are shown in the Figure \ref{fig:trifingerRealRobot}. It is worth noting that the robot is situated in a different location and therefore the sim-to-real transfer was done remotely.

\subsubsection{Allegro Hand}

We learn cube orientation with Allegro Hand and use the same reward as for the Shadow Hand as well similar observation scheme, with the only difference --- smaller number of observations because of the different number of fingers in Allegro Hand --- that it has 4 fingers instead of 5 and fewer degrees of freedom as a result, shown in Appendix~\ref{sec:env-details:allegro}.

Figure \ref{fig:hands-rewards}(d) shows the reward curves for Allegro Hand and Figure \ref{fig:hands:consecutive-success}(d) shows consecutive successes achieved. Interestingly, despite having fewer degrees of freedom this hand does not achieve as high consecutive successes as Shadow hand. This is because the wrist is fixed and fingers are slightly longer. We observed in Shadow hand experiment that having a movable wrist allows for better manipulation when reorienting the cube.

\section{Summary}
\label{sec:summary}
We show that Isaac Gym is a high performance and high-fidelity framework that allows blistering fast training on many challenging simulated robotic environments on a single NVIDIA A100 GPU that previously would have required large heterogeneous clusters of CPUs and GPUs using a conventional RL setup with CPU-only simulators. Moreover, the simulation backend \cite{nvidia-physx} is also suited for learning contact-rich manipulations as confirmed by our sim-to-real transfer demonstrations with ANYmal locomotion and TriFinger cube reposing. 

\section{Acknowledgements}

We would like to thank the following for  additional hard work helping us with this work.

Jonah Alben, Rika Antonova, Ayon Bakshi, Dennis Da, Shoubhik Debnath, Clemens Eppner, Dieter Fox, Animesh Garg, Renato Gasoto, Isabella Huang, Andrew Kondrich, Rev Lebaredian, Qiyang Li, Jacky Liang, Denys Makoviichuk, Brendon Matusch, Hammad Mazhar, Mayank Mittal, Adam Moravansky, Yashraj Narang, Oyindamola Omotuyi, Fabio Ramos, Andrew Reidmeyer, Philipp Reist, Tony Scudiero, Mike Skolones, Balakumar Sundaralingam, Liila Torabi, Cameron Upright, Zhaoming Xie, Winnie Xu, Yuke Zhu, and the rest of the NVIDIA PhysX, Omniverse, and robotics research teams. We also thank Jason Peng and Josiah Wong for the help in AMP and Franka Cube Stacking experiments.

Thanks are also due to open-source community projects like Matplotib\cite{Hunter:2007}, Python\cite{Python:etal:2009}, NumPy\cite{Numpy:etal:2020}, PyTorch\cite{PyTorch:etal:2019}, Tensorboard\cite{Tensorflow2015-whitepaper}, Tensorboard Aggregator\cite{tb-aggregator} and SciencePlots\cite{SciencePlots} which we used heavily in this work. We are thankful to Overleaf \cite{overleaf} for hosting our latex project.

\clearpage
\newpage
\bibliographystyle{unsrtnat}
\bibliography{main_arXiv_2021}

\newpage

\appendix

\section{Appendix}

\subsection{Tendons}
\label{sec:tendon-details}

We simulate tendons as part of the Shadow Hand environment and describe the details of this simulation here.

\subsubsection{Fixed Tendons}

Fixed tendons are an abstract mechanism that couple degrees of freedom (DOF) of an articulation. A fixed tendon is composed of a tree of tendon joints, where each joint is associated with exactly one axis of a link’s incoming articulation joint. In the following, when we refer to a tendon joint's position, we mean the position of the axis of this associated articulation joint. 

In addition, each tendon joint has a coefficient that determines the contribution of the (rotational or translational) joint position to the length of the tendon, which is evaluated recursively by traversing the tree: The length at a given tendon joint is the length at its parent tendon joint plus its joint position scaled by the coefficient.

Given the tendon length at each joint, the tendon applies a spring force (or torque) to the joint's child link that is proportional to the deviation of the tendon length from a desired (tendon-wide) rest length. An equal and opposing force is applied to the parent link of the root tendon joint; conceptually, each tendon joint is a virtual joint drive between the root parent link and the tendon joint's child link. 
In addition to the spring force, the tendon-joint applies a damping force that is proportional to and acting against the velocity of the virtual root-to-child link joint.

Analogous to the length dynamics, the tendon supports length limits that apply an additional force or torque that is proportional to the deviation from set limits.

\subsubsection{Spatial Tendons}

Spatial tendons create line-of-sight distance constraints between links of a single articulation. In particular, spatial tendons run through attachments that are positioned relative to an articulation link, and their length is defined as a weighted sum of the distance between the attachments in the tendon. It is possible to create multiple attachments per link, for example for tendon-routing purposes. In contrast to fixed tendons, spatial tendons are not constrained to follow the articulation topology.

Same as fixed tendons, spatial tendons may branch, in which case the tendon splits up into multiple conceptual sub-tendons, one for each root-to-leaf path in the tendon tree. Length and limit constraints are evaluated per sub-tendon, and have spring-damper dynamics that may both contract and extend the tendon (one may use appropriately set limits to achieve a one-sided, string-like constraint).

The sub-tendon constraint force acts on the leaf and root attachments, in the direction of its parent for the leaf, and in the direction of the child on the path to the sub-tendon leaf for the root. However, the force does not propagate further and act on any intermediate attachments between root and leaf.

\subsection{Observations \& Rewards}
\label{sec:env-details}

In this section we describe the reward and observations for each environment in detail.

\subsubsection{Ant and Humanoid environments}
\label{sec:env-details:mujoco}

Both the Ant and Humanoid environments use the same reward formulation, namely:

\begin{align*}
    \mathsf{R} &= \mathsf{R_{progress}} + \mathsf{R_{alive}}\times \mathbbm{1}{(\mathsf{torso\_height} \ge \mathsf{termination\_height})} + \mathsf{R_{upright}} \\ 
         &+ \mathsf{R_{heading}} + \mathsf{R_{effort}} + \mathsf{R_{act}} + \mathsf{R_{dof}}\\
         &+ \mathsf{R_{death}} 
         \times  \mathbbm{1}{(\mathsf{torso\_height} \leq \mathsf{termination\_height})}
\end{align*}

where

\begin{align*}
\mathsf{R_{progress}} &= \mathsf{potential} - \mathsf{prev\_potential} \\ 
\mathsf{R_{upright}} &= \mathsf{dot(torso\_up\_vector, up\_vector)} > 0.93\\ %
\mathsf{R_{heading}} &= \mathsf{heading\_weight} *
\begin{cases}
    1.0, & \text{if } \mathsf{norm\_angle\_to\_target}\geq 0.8\\
    \frac{\mathsf{norm\_angle\_to\_target}}{0.8}, & \text{otherwise}
\end{cases}\\ 
\mathsf{R_{act}} &= - \sum ||\mathsf{actions}||^2 \\
\mathsf{R_{effort}} &= \sum_{i=1}^{N} \mathsf{actions}_{i} \times \mathsf{normalized\_motor\_strength}_{i} \times \mathsf{dof\_velocity}_{i} \\
\mathsf{potential} &= -\frac{||\mathsf{p_{target}} - \mathsf{p_{torso}}||_{2}}{dt}
\end{align*}

\paragraph{Ant}
\label{sec:env-details:ant}

Reward described in Section~\ref{sec:env-details:mujoco}. Observations detailed in Table~\ref{tab:env-details:ant:obs}.

\begin{table}[hbt!]
\begin{minipage}{.45\linewidth}
\centering
\resizebox{\linewidth}{!}{%
    \centering
    \begin{tabular}{|l|l|c|} 
    \hline
    \rowcolor{Light}\multicolumn{2}{|c|}{Observation space}                                   & Degrees of freedom  \\ 
    \hline
    \multicolumn{2}{|l|}{Torso vertical position}                    & 1                  \\ 
    \hline
    \multirow{2}{*}{Velocity}                    & positional         & 3                  \\ 
                                                 & angular            & 3                  \\ 
    \hline
    \multicolumn{2}{|l|}{Yaw, roll, angle to target}                    & 3                  \\ 
    \hline
    \multicolumn{2}{|l|}{Up and heading vector proj.}                    & 2                  \\ 
    \hline
    \multirow{2}{*}{DOF measurements}                    & position         & 8                  \\ 
                                                 & velocity            & 8                  \\ 
    \hline
    \multicolumn{2}{|l|}{Sensor forces, torques}                    & 24                  \\ 
    \hline
    \multicolumn{2}{|l|}{Actions}                    & 8                  \\ 
    \hline
    \rowcolor{Light} \multicolumn{2}{|c|}{Total number of observations}                                  & 60                 \\
    \hline
    \end{tabular}
    }
    \caption{Observations used for Ant training.}
    \label{tab:env-details:ant:obs}
\end{minipage}%
\hfill
\begin{minipage}{.45\linewidth}
\resizebox{\linewidth}{!}{%
    \centering
    \begin{tabular}{|l|l|c|} 
    \hline
    \rowcolor{Light}\multicolumn{2}{|c|}{Observation space}                                   & Degrees of freedom  \\ 
    \hline
    \multicolumn{2}{|l|}{Torso vertical position}                    & 1                  \\ 
    \hline
    \multirow{2}{*}{Velocity}                    & positional         & 3                  \\ 
                                                 & angular            & 3                  \\ 
    \hline
    \multicolumn{2}{|l|}{Yaw, roll, angle to target}                    & 3                  \\ 
    \hline
    \multicolumn{2}{|l|}{Up and heading vector proj.}                    & 2                  \\ 
    \hline
    \multirow{2}{*}{DOF measurements}                    & position         & 21                  \\ 
                                                 & velocity            & 21                  \\ 
                                                 & force            & 21                  \\ 
    \hline
    \multicolumn{2}{|l|}{Sensor forces/torques}                    & 12                  \\ 
    \hline
    \multicolumn{2}{|l|}{Actions}                    & 21                  \\ 
    \hline
    \rowcolor{Light} \multicolumn{2}{|c|}{Total number of observations}                                  & 108                 \\
    \hline
    \end{tabular}
    }
    \caption{Observations used for Humanoid training.}
    \label{tab:env-details:humanoid:obs}
\end{minipage}
\end{table}

\paragraph{Humanoid}
\label{sec:env-details:humanoid}

Reward described in Section~\ref{sec:env-details:mujoco}.  Observations detailed in Table~\ref{tab:env-details:humanoid:obs}.

\subsubsection{Locomotion environments}
\label{sec:env-details:locomotion}

\paragraph{Ingenuity}
\label{sec:env-details:ingenuity}

Observations detailed in Table~\ref{tab:env-details:ingenuity:obs}. The reward function is as follows:

\begin{align*}
\mathsf{R} &= \mathsf{R_{pos}} \times (1 + \mathsf{R_{upright}} + \mathsf{R_{spin}})
\end{align*}

\textbf{reaching cost} 
\begin{equation*}
\mathsf{R_{pos}} = \frac{1}{1 + ||\mathsf{dist\_to\_target}||^2}
\end{equation*}

\textbf{spinning cost} 
\begin{equation*}
\mathsf{R_{spin}} = \frac{1}{1 + ||\mathsf{spin\_rate}||^2}
\end{equation*}

\textbf{upright cost} 
\begin{equation*}
\mathsf{R_{upright}} = \frac{1}{1 + \mathsf{local\_up\_vector}_z^2}
\end{equation*}

\begin{table}[hbt!]

\begin{minipage}{.45\linewidth}
\centering
\resizebox{\linewidth}{!}{%

    \centering
    \begin{tabular}{|l|l|c|} 
    \hline
    \rowcolor{Light}\multicolumn{2}{|c|}{Observation space}                                   & Degrees of freedom  \\ 
    \hline
    \multicolumn{2}{|l|}{Offset from target}                    & 3                  \\ 
    \hline
    \multicolumn{2}{|l|}{Rotation}                    & 4                  \\ 
    \hline
    \multirow{2}{*}{Velocity}                    & positional         & 3                  \\ 
                                                 & angular            & 3                  \\ 
    \hline
    \rowcolor{Light} \multicolumn{2}{|c|}{Total number of observations}                                  & 13                 \\
    \hline
    \end{tabular}
    }
    \caption{Observations used for ingenuity training.}
    \label{tab:env-details:ingenuity:obs}
\end{minipage}
\begin{minipage}{.45\linewidth}
\centering
\resizebox{\linewidth}{!}{%
    \begin{tabular}{|l|l|c|} 
    \hline
    \rowcolor{Light}\multicolumn{2}{|c|}{Observation space}                                   & Degrees of freedom  \\ 
    \hline
    \multirow{2}{*}{Base velocity} & positional & 3 \\ 
    & angular & 3 \\ 
    \hline
    \multicolumn{2}{|l|}{Body-relative gravity} & 3 \\
    \hline
    \multicolumn{2}{|l|}{Target X, Y, yaw velocities} & 3 \\ 
    \hline
    \multirow{2}{*}{DOF states} & position & 12 \\ 
     & velocity & 12 \\ 
    \hline
    \multicolumn{2}{|l|}{Actions} & 12 \\ 
    \hline
    \rowcolor{Light} \multicolumn{2}{|c|}{Total number of observations}              & 48 \\
    \hline
    \end{tabular}
    }
    \caption{Observations used for ANYmal training.}
    \label{tab:env-details:anymal:obs}
\end{minipage}
\end{table}

\paragraph{ANYmal Locomotion}
\label{sec:env-details:anymal-locomotion}

For the included flat-terrain environment, observations are detailed in Table~\ref{tab:env-details:anymal:obs} and the reward function is as follows:

\begin{align*}
\mathsf{R} &= c_1\mathsf{R_{vel, xy}} + c_2\mathsf{R_{vel, yaw}} + c_3\mathsf{R_{torque}}\\
\end{align*}
Reward terms are defined in Table \ref{table:anymal_rewards} and symbols in Table \ref{table:anymal_nomenclature}.

For rough terrain locomotion with sim-to-real, we extend the observations with 140 terrain heights around the robot's base and use the more complex reward function:
\begin{align*}
\mathsf{R} = &c_1\mathsf{R_{vel, xy}} + c_2\mathsf{R_{vel, yaw}} + c_3\mathsf{R_{vel, z}} + c_4\mathsf{R_{vel, pitch/roll}} + c_5\mathsf{R_{joint\, vel/acc}} +\\ &c_6\mathsf{R_{torque}} + c_7\mathsf{R_{rate}} + c_8\mathsf{R_{collision}} + c_9\mathsf{R_{airtime}}
\end{align*}

\begin{table}[H]\centering
\begin{minipage}{.35\linewidth}
\centering
\resizebox{\linewidth}{!}{%
\begin{tabular}{|l|c|}
    \hline
    \rowcolor{Light} Quantity & Symbol\\
    \hline
     Joint positions & $\mathbf{q}_j$\\
     \hline
     Joint velocities & $\dot{\mathbf{q}_j}$\\
     \hline
     Joint accelerations & $\ddot{\mathbf{q}_j}$\\
     \hline
     Target joint positions & $\ddot{\mathbf{q}^*_j}$\\
     \hline
     Joint torques & $\boldsymbol{\tau}_j$\\
     \hline
     Base linear velocity & $\mathbf{v}_{b}$\\
     \hline
     Base angular velocity & $\boldsymbol{\omega}_{b}$\\
     \hline
     Commanded base linear velocity & $\mathbf{v}^*_{b}$\\
     \hline
     Commanded base angular velocity & $\boldsymbol{\omega}^*_{b}$\\
     \hline
     Number of collisions & $n_{c}$\\
     \hline
     Feet air time & $\mathbf{t}_{air}$\\
     \hline
     Environment time step & $dt$\\
     \hline
\end{tabular}
}
\caption{Definition of symbols.}
\label{table:anymal_nomenclature}
\end{minipage}%
\hfill
\begin{minipage}{.6\linewidth}
\centering
\resizebox{\linewidth}{!}{%
\begin{tabular}{|l|c|c|c|}
    \hline
     \rowcolor{Light} Reward & Symbol & Definition & Weight\\
     Linear velocity tracking & $\mathsf{R_{vel, xy}}$ & $\phi(\mathbf{v}^*_{b,xy} - \mathbf{v}_{b,xy})$ & $1 dt$ \\
     \hline
     Angular velocity tracking & $\mathsf{R_{vel, yaw}}$ & $\phi(\boldsymbol{\omega}^*_{b,z} - \boldsymbol{\omega}_{b,z})$ & $0.5 dt$\\
     \hline
     Linear velocity penalty & $\mathsf{R_{vel, z}}$ & $-\mathbf{v}_{b,z}^2$ & $4 dt$ \\
     \hline
     Angular velocity penalty & $\mathsf{R_{vel, pitch/roll}}$ & $-||\boldsymbol{\omega}_{b,xy}||^2$ & $0.05 dt$\\
     \hline
     Joint motion & $\mathsf{R_{joint\,vel/acc}}$ & $-||\ddot{\mathbf{q}_j}||^2 - ||\dot{\mathbf{q}_j}||^2$ & $0.001 dt$ \\
     \hline
     Joint torques & $\mathsf{R_{torque}}$ & $-||\boldsymbol{\tau}_j||^2$ & $0.00002 dt$\\
     \hline
     Action rate & $\mathsf{R_{rate}}$ & $-||\dot{\mathbf{q}^*_j}||^2$ & $0.25 dt$ \\
     \hline
     Collisions & $\mathsf{R_{coll.}}$ & $-n_{collision}$ & $0.001 dt$\\
     \hline
     Feet air time & $\mathsf{R_{air time}}$ & $ \sum_{f=0}^{4}(\mathbf{t}_{air, f} - 0.5)$ & $2 dt$\\
    \hline
\end{tabular}
}
\caption{Definition of reward terms, with $\phi(x):=\exp(-\frac{||x||^2}{0.25})$. The z axis is aligned with gravity.}
\label{table:anymal_rewards}
\end{minipage}
\end{table}

\paragraph{Adversarial Imitation Learning}
\label{sec:env-details:amp}
AMP learns an imitation objective using an adversarial discriminator $D$, trained to differentiate between motion from the dataset $\mathcal{M}$ and motions produced by the policy $\pi$,
\begin{align*}
    \mathop{\mathrm{arg \ min}}_{D} \quad -\mathbb{E}_{s, s' \sim p_\mathcal{M}(s, s')} \left[ \mathrm{log} D(s, s') \right] - \mathbb{E}_{s, s' \sim p_\pi(s, s')} \left[ \mathrm{log}\left(1 - D(s, s') \right) \right],
\end{align*}
where $p_\mathcal{M}(s, s')$ denotes the likelihood of observing a state transition from $s$ to $s'$ in the motion data, and $p_\pi(s, s')$ is likelihood of a state transition under the policy. The discriminator can then be used to specify rewards $r_t$ for training a policy to imitate behaviors shown in the motion data
\begin{align*}
    r_t = - \mathrm{log} \left(1 - D(s_t, s_{t+1}) \right).
\end{align*}
This objective, in effect encourages the policy to produce behaviors that fool the discriminator into classifying them as behaviors from the reference motion data. 

\begin{table}[hbt!]

\begin{minipage}{.45\linewidth}
\centering
\resizebox{\linewidth}{!}{%

    \begin{tabular}{|l|l|c|} 
    \hline
    \rowcolor{Light}\multicolumn{2}{|c|}{Observation space}     & Degrees of freedom  \\ 
    \hline
    \multicolumn{2}{|l|}{Pelvis vertical height}                    & 1                  \\ 
    \hline
    \multicolumn{2}{|l|}{Pelvis rotation}                    & 6                  \\ 
    \hline
    \multirow{2}{*}{Pelvis Velocity}                    & positional         & 3                  \\ 
                                                 & angular            & 3                  \\
    \hline
    \multirow{2}{*}{DOF measurements}                    & position         & 52                  \\ 
                                                 & velocity            & 28                  \\ 
    \hline
    \multicolumn{2}{|l|}{Key point position}                    & 15 \\ 
    \hline
    \rowcolor{Light} \multicolumn{2}{|c|}{Total number of observations}                                  & 108                 \\
    \hline
    \end{tabular}
    }
    \caption{Observations used for AMP training with a humanoid character. 3D rotations are represented using a 6D tangent-normal encoding.}
    \label{tab:env-details:amp:obs}
\end{minipage}%
\hfill
\begin{minipage}{.45\linewidth}
 \centering
\resizebox{\linewidth}{!}{%
    \begin{tabular}{|l|l|c|} 
    \hline
    \rowcolor{Light}\multicolumn{2}{|c|}{Observation space} & Degrees of freedom  \\ 
    \hline
    \multirow{2}{*}{Joint DOFs} & arm position & 7 (Joint Torque only) \\ 
     & eef position  & 2 \\ 
    \hline
    \multicolumn{2}{|l|}{EEF pose} & 7  \\
    \hline
    \multicolumn{2}{|l|}{Cube A pose} & 7  \\
    \hline
    \multicolumn{2}{|l|}{Cube A to Cube B position} & 3  \\
    \hline
    \rowcolor{Light} \multicolumn{2}{|c|}{Total number of observations} & 19 / 26                 \\
    \hline
    \end{tabular}
    }
    \caption{Franka Cube Stack observations. Note that pose observations include the global 3-dim cartesian position and 4-dim quaternion orientation, and the arm joint position observations are only provided if using joint torque control.}
    \label{tab:env-details:franka-cube-stack:obs}
\end{minipage}
\end{table}

\paragraph{Franka Cube Stack}
\label{sec:env-details:franka-cube-stack}

Observations are detailed in Table~\ref{tab:env-details:franka-cube-stack:obs}. The reward function used is as follows:

\begin{align*}
    \mathsf{R} = \text{max}\left( \mathsf{R_{stack}}, \mathsf{R_{align}} + \mathsf{R_{lift}} + \mathsf{R_{reach}} \right)
\end{align*}

where:

\begin{align*}
    \mathsf{R_{stack}} = \quad& \mathsf{w_{stack}} \times \mathbbm{1}((\mathsf{height_{cubeA}} > \mathsf{height_{cubeB}}) \& (\mathsf{cubeA\_aligned\_cubeB}) \& (\mathsf{gripper\_away\_from\_cubeA})), \\
    \mathsf{R_{align}} = \quad& \mathsf{w_{align}} \times (1 - \tanh(10 \times \mathsf{cubeA\_to\_B\_xy\_dist})) \times \mathbbm{1}(\mathsf{cubeA\_is\_lifted}), \\
    \mathsf{R_{lift}} = \quad& \mathsf{w_{lift}} \times \mathbbm{1}(\mathsf{cubeA\_is\_lifted}), \\
    \mathsf{R_{reach}} = \quad& \mathsf{w_{reach}} \times \bigg(1 - \tanh\big(\frac{10}{3} \times (\mathsf{dist(cubeA, gripper)} + \mathsf{dist(cubeA, lfinger)} + \mathsf{dist(cubeA, rfinger)})\big)\bigg)
\end{align*}

We set $\mathsf{w_{stack}} = 16.0$, $\mathsf{w_{align}} = 2.0$, $\mathsf{w_{lift}} = 1.5$, and $\mathsf{w_{reach}} = 0.1$

\subsubsection{Robotic Hands}
\label{sec:robotic-hands-rewards}

\paragraph{Shadow Hand}
\label{sec:env-details:shadow-hand}

The reward function for Shadow Hand is as follows:

\begin{equation*}
\mathsf{R} = \mathsf{w_{dist}}\mathsf{R_{dist}} + \mathsf{R_{rot}} + \mathsf{w_{act}}\mathsf{R_{act}}
\end{equation*}

\textbf{distance cost} 
\begin{equation*}
\mathsf{R_{dist}} = ||\mathsf{p_{obj}} - \mathsf{p_{target}}||_2
\end{equation*}

\textbf{orientation cost} 
\begin{equation*}
\mathsf{rot\_dist} = 2 \times \arcsin (\max (1, ||\mathsf{q_{obj}} * \overline{\mathsf{q_{target}}}||_2))
\end{equation*}

\begin{equation*}
\mathsf{R_{rot}} = \frac{1}{\mathsf{|rot\_dist|}+0.1}
\end{equation*}

\textbf{action smoothness cost} 
\begin{equation*}
\mathsf{R_{act}} = \sum ||\mathsf{actions}||^2
\end{equation*}

where $\mathsf{w_{dist}} = -10$ and $\mathsf{w_{act}} = -2e-4$.

\begin{table}[hbt!]
\begin{minipage}{.45\linewidth}
\centering
\resizebox{\linewidth}{!}{%
\begin{tabular}{|l|l|c|} 
\hline
\rowcolor{Light}\multicolumn{2}{|c|}{Observation space}                                   & Degrees of freedom  \\ 
\hline
\multirow{3}{*}{Finger joints}                    & position         & 24                  \\ 
                                                       & velocity  & 24                  \\ 
                                                       & force            & 24                  \\ 
\hline
\multirow{4}{*}{Cube pose}                           & translation      & 3                   \\ 
                                                       & quaternion~      & 4                   \\ 
                                                       & linear velocity  & 3                   \\ 
                                                       & angular velocity & 3                   \\ 
\hline
Cube rotation relative to goal                                      & quaternion       & 4                   \\ 
\hline
\multirow{2}{*}{Goal pose}                             & translation      & 3                   \\ 
                                                       & quaternion       & 4                   \\ 
\hline                                                       
\multirow{6}{*}{5 $\times$ Finger tips} & position                    & 3                                        \\
                                        & quaternion                  & 4                                        \\
                                        & linear velocity             & 3                                        \\
                                        & angular velocity            & 3                                        \\ 
                                        & force                       & 3                                        \\
                                        & torque                      & 3                                        \\ 
\hline
\multicolumn{2}{|c|}{Previous action output from policy}                                  & 20  \\
\hline
\rowcolor{Light} \multicolumn{2}{|c|}{Total number of observations}                                  & 211                 \\
\hline
\end{tabular}
}
\caption{Observations for the Shadow Hand \textbf{Standard} environment.}
\label{tab:env-details:shadow-standard:obs}
\end{minipage}%
\hfill
\begin{minipage}{.45\linewidth}
\centering
\resizebox{\linewidth}{!}{%

\begin{tabular}{|l|l|c|} 
\hline
\rowcolor{Light}\multicolumn{2}{|c|}{Observation space}                                   & Degrees of freedom  \\ 
\hline
\multirow{1}{*}{5 $\times$ Finger joints} & position                    & 3                                        \\
\hline
\multirow{1}{*}{Cube pose}                           & translation      & 3                   \\ \hline
Cube rotation relative to goal                           & quaternion       & 4                   \\ 
\hline
\multicolumn{2}{|c|}{Previous action output from policy}        & 20  \\
\hline
\rowcolor{Light} \multicolumn{2}{|c|}{Total number of observations}                                  & 42                 \\
\hline
\end{tabular}
}
\caption{Observations for the Shadow Hand \textbf{OpenAI} environment. The observations of the critic are the same as for Shadow Hand Standard (see Table~\ref{tab:env-details:shadow-standard:obs}).}
\label{tab:env-details:shadow-openai:obs}
\end{minipage}
\end{table}

There are two different variants of observations used. In the Shadow Hand \textbf{Standard} environment, the observations are as shown in Table~\ref{tab:env-details:shadow-standard:obs}. In The ShadowHand \textbf{OpenAI} environment, in order to compare to compare to \citep{openai-sh}, we use observations as shown in Table~\ref{tab:env-details:shadow-openai:obs}. Further details of the Shadow Hand environments are available in Appendix~\ref{sec:shadow-hand-details}. Below we provide the code snippet to compute the reward as used in our implementation.

\begin{lstlisting}[language=Python, title={Reward function for cube orientation for Shadow Hand experiments.} ,label={lst:shadow_hand_reward}, captionpos=b]
@torch.jit.script
def compute_hand_reward(
    object_pos, object_rot, target_pos, target_rot, actions, 
    dist_reward_scale: float, rot_reward_scale: float, rot_eps: float,
    action_penalty_scale: float, success_tolerance: float, reach_goal_bonus: float, 
    fall_dist: float,fall_penalty: float):
    
    #dist_reward_scale: -10.0
    #rot_reward_scale: 1.0
    #rot_eps: 0.1
    #action_penalty_scale: -0.0002
    #reach_goal_bonus: 250
    #fall_distance: 0.24
    #fall_penalty: 0.0
    
    # Distance from the hand to the object
    goal_dist = torch.norm(object_pos - target_pos, p=2, dim=-1)

    # Orientation alignment for the cube in hand and goal cube
    quat_diff = quat_mul(object_rot, quat_conjugate(target_rot))
    rot_dist  = 2.0 * torch.asin(torch.clamp(torch.norm(quat_diff[:, 0:3], p=2, dim=-1), max=1.0))
    
    # Orientation reward
    rot_rew = 1.0/(torch.abs(rot_dist) + rot_eps)
    
    # action smoothness reward 
    action_penalty = torch.sum(actions ** 2, dim=-1)

    # Total reward is: position distance + orientation alignment + action regularization + success bonus + fall penalty
    reward = goal_dist * dist_reward_scale + rot_rew * rot_reward_scale 
           + action_penalty * action_penalty_scale

    # Find out which envs hit the goal and update successes count
    goal_resets = torch.where(torch.abs(rot_dist) <= success_tolerance, 
                              torch.ones_like(reset_goal_buf), reset_goal_buf)

    # Success bonus: orientation is inside the `success_tolerance` of goal orientation
    reward = torch.where(goal_resets == 1, reward + reach_goal_bonus, reward)

    # Fall penalty: distance to the goal is larger than a threashold
    reward = torch.where(goal_dist >= fall_dist, reward + fall_penalty, reward)

    return reward
\end{lstlisting}

\paragraph{Trifinger}
\label{sec:env-details:trifinger}

\begin{table}[hbt!]
\begin{minipage}{.45\linewidth}
\centering
\resizebox{\linewidth}{!}{%
\begin{tabular}{|l|l|c|} 
\hline
\rowcolor{Light}\multicolumn{2}{|c|}{Observation space}                                   & Degrees of freedom  \\ 
\hline
\multirow{2}{*}{Finger joints}                    & position         & 9                  \\ 
                                                       & velocity  & 9                  \\ 
\hline
\multirow{2}{*}{Cube pose}                           & translation      & 3                   \\ 
                                                       & quaternion~      & 4                   \\ 
\hline
\multirow{2}{*}{Goal pose}                           & translation      & 3                   \\ 
                                                       & quaternion~      & 4                   \\ 
\hline
\multicolumn{2}{|c|}{Previous action output from policy}                                  & 9  \\
\hline
\rowcolor{Light} \multicolumn{2}{|c|}{Total number of observations}                                  & 41                 \\
\hline
\end{tabular}
}
\caption{Trifinger Actor Observations.}
\label{tab:env-details:trifinger:obs}
\end{minipage}%
\hfill
\begin{minipage}{.45\linewidth}
\centering
\resizebox{\linewidth}{!}{%

\begin{tabular}{|l|l|c|} 
\hline
\rowcolor{Light}\multicolumn{2}{|c|}{Observation space}                                   & Degrees of freedom  \\ 
\hline
\multicolumn{2}{|c|}{Actor Observations (see Table~\ref{tab:env-details:trifinger:obs})}                                  & 41  \\
\hline
\multirow{2}{*}{Cube pose}                           & linear velocity      & 3                   \\ 
                                                       & angular velocity      & 3                   \\ 

\hline
\multirow{6}{*}{3 $\times$ Finger tips} & position                    & 3                                        \\
                                        & quaternion                  & 4                                        \\
                                        & linear velocity             & 3                                        \\
                                        & angular velocity            & 3                                        \\ 
                                        & force                       & 3                                        \\
                                        & torque                      & 3                                        \\

\hline
\multirow{1}{*}{Finger joints}                    & terque         & 9                  \\ 
\hline
\rowcolor{Light} \multicolumn{2}{|c|}{Total number of observations}                                  & 113                 \\
\hline
\end{tabular}
}
\caption{Trifinger Critic Observations}
\label{tab:env-details:trifinger:critic-obs}
\end{minipage}
\end{table}

Our total reward is defined as:

\begin{align*}
\mathsf{R} = \mathsf{w_{og}} \mathsf{R_{object\_goal}} 
+ \mathsf{w_{fo}} \mathsf{R_{fingertip\_object}} \times \mathbbm{1}(\mathsf{timesteps} \leq 5e7) 
+ \mathsf{w_{fv}} \mathsf{R_{fingertip\_velocity}} 
\end{align*}

\textbf{reposing cost} 
\begin{equation*}
\mathsf{R_{object\_goal}} = \mathcal{K}(||\mathsf{t_{curr}} - \mathsf{t_{target}}||_2) + \frac{1}{3 \times \mathsf{|rot\_dist|}+0.01}
\end{equation*}

\textbf{fingertips interaction cost} 
\begin{equation*}
\mathsf{R_{fingertip\_object}} = \sum_{i \in \mathsf{fingertips}} \Delta^t_{i}
\end{equation*}

\textbf{fingertips smoothness cost} 
\begin{equation*}
\mathsf{R_{fingertip\_velocity}} = \sum_{i \in \mathsf{fingertips}} ||\mathsf{fingertip\_speed}_i||^2 
\end{equation*}

$\Delta^t_i$ denotes the change across the timestep of the fingertip distance to the centroid of the object and was found to be helpful in \citep{causalworld}. Formally, $\Delta^t_i=||\mathsf{ft_{i, t}}-\mathsf{t_{curr, t}}||_2 - ||\mathsf{ft_{i, t-1}}-\mathsf{t_{curr, t-1}}||_2$, where $\mathsf{t_{curr, t}}$ is position of the cube centroid and $\mathsf{ft_i}$ denotes the position of the $i$-th fingertip at time $t$. 

$\mathsf{rot\_dist}$ is the angluar difference between the current and target cube pose, $\mathsf{rot\_dist} = 2 \times \arcsin (\min(1.0, ||\mathsf{q_{diff}}||_2)), \mathsf{q_{diff}} = \mathsf{q_{curr}} \mathsf{q_{target}^{*}}$.
Following \citep{Hwangbo_2019}, a logistic kernel is used to convert tracking error in euclidean space into a bounded reward function, with $\mathcal{K}(x) = \left(e^{ax} + b + e^{-ax}\right)^{-1}$, where $a$ is a scaling factor; we use $a=50$. See \citep{isaacgym-trifinger} for a more thorough motivation and description of these reward terms.

\paragraph{Allegro}
\label{sec:env-details:allegro}

The reward formulation is identical to that used in Shadow Hand - see Appendix~\ref{sec:env-details:shadow-hand}. The observations are also identical, save for the change in number of fingers.

\subsection{Hyperparameters for Training PPO}

\begin{table}[h]
\resizebox{\linewidth}{!}{%
\begin{tabular}{|c|c|c|c|c|c|c|c|}
\hline
\rowcolor{Light} %
Environment    & \# Environments & \multicolumn{1}{l|}{\cellcolor{Light}KL Threshold} & Mini-batch Size & \multicolumn{1}{l|}{\cellcolor{Light} Horizon Length} & \# PPO Epochs & Hidden Units   &   Training Steps  \\ \hline
Ant            & 4096            & 8e-3                                                      & 32768           & 16                                                          & 4         & 256, 128, 64 & 32M  \\ \hline
Humanoid       & 4096            & 8e-3                                                      & 32768           & 32                                                          & 5         & 400, 200, 100 & 327M \\ \hline
Ingenuity      & 4096            & 1.6 e-2                                                   & 32768           & 16                                                          & 8         & 256, 256, 128 & 32M    \\ \hline
ANYmal         & 8192            & 1e-2                                                      & 32768           & 16                                                          & 5         & 256, 128, 64 & 65M  \\ \hline
ANYmal Terrain & 4096            & 1e-2                                                      & 24576           & 24                                                          & 10        & 512, 256, 128 & 150M  \\ \hline
AMP            & 4096            & 2e-1                                             & 16384             & 32                                                   &              8   & 1024, 512 & 39M  \\ \hline
Franka         & 16384           & 1.6 e-2                                                   & 131072          & 32                                                          & 4         & 256, 128, 64  & 786M   \\ \hline
SH Standard    & 16384           & 1.6 e-2                                                   & 32768           & 8                                                           & 5         & 512, 512, 256, 128 & 655M \\ \hline
SH OpenAI  FF    & 16384           & 1.6 e-2                                                   & 32768           & 8                                                           & 5         & 400, 400, 200, 100 & 655M\\ \cline{7-7} 
\hline
SH OpenAI LSTM    & 8192           & 1.6 e-2                                                   & 32768           & 16                                                           & 4         & lstm: 1024, mlp: 512  & 925M\\ \cline{7-7} 
\hline
\end{tabular}
}
\vspace{2mm}
    \caption{Hyperparameters used for training in each environment. Allegro shares the parameters for Shadow Hand. The hidden units are ELU for every environment except AMP, where ReLU units are used. Additionally, every environment uses an adaptive learning rate with a KL divergence target specified in the \textit{KL Threshold} column, except for AMP which uses a fixed learning rate of 2e-5 and fixed KL theshold of 2e-1. The SH OpenAI LSTM experiment uses an LSTM layer of 1024 hidden dims followed by MLP of 512 dims, and a fixed learning rate of 1e-4 for the value function.}
    \label{tab:hyperparameters}

\end{table}

\subsection{Shadow Hand Details}
\label{sec:shadow-hand-details}

As mentioned previously, we implemented two variants of the Shadow Hand environment. The \textbf{Standard} variant uses privileged policy observations and no Domain Randomization, in order to provide a quick training example to test Reinforcement Learning algorithms on. The \textbf{OpenAI} variant uses asymmetric observations, such that it would be possible to transfer the policy to the real world, mimicing the setup in \citep{openai-sh}.

\subsubsection{Randomizations}
\label{sec:shadow-appendix:randomizations}

Isaac Gym implements a high-level API that simplifies setting up physics domain randomization parameters and schedule in yaml configuration files and is very extensible. Here we detail the randomization parameters that we used.

\paragraph{Random forces on the object.}
Following \citep{openai-sh} unmodeled dynamics is represented by applying random forces on the object. The probability $p$ that a random force is applied is sampled at the beginning of the randomization episode from the loguniform distribution between $0.1\%$ and $10\%$. Then, at every timestep, with probability $p$ we apply a random force from the $3$-dimensional Gaussian distribution with the standard deviation equal to $1~m/s^2$ times the mass of the object on each coordinate and decay the force with the coefficient of $0.99$ per $50$ms.

\paragraph{Physics randomizations.} Physical parameters like friction, link and object masses, cube size, joint and tendon properties, as well as correlated noise parameters are randomized every time an environment is reset, with a minimum interval of 720 steps. \autoref{table:rand-physics} lists all physics parameters that are randomized.

\begin{table}
    \footnotesize
    \centering
    \renewcommand{\arraystretch}{1.3}
    \begin{tabular}{@{}lll@{}}
        \toprule
        \rowcolor{Light}\textbf{Parameter} & \textbf{Scaling factor range} & \textbf{Additive term range} \\ \midrule
        object dimensions & $\mbox{uniform}([0.95,1.05])$ & \\
        object and robot link masses & $\mbox{uniform}([0.5,1.5])$ & \\
        surface friction coefficients & $\mbox{uniform}([0.7,1.3])$ & \\
        robot joint damping coefficients & $\mbox{loguniform}([0.3,3.0])$ & \\
        actuator force gains (P term) & $\mbox{loguniform}([0.75,1.5])$ & \\ \hline
        joint limits & & $\mathcal{N}(0,0.15)~\si{\radian}$  \\
        gravity vector (each coordinate) && $\mathcal{N}(0,0.4)~\si{\m\per\s^2}$ \\ %
        \bottomrule
    \end{tabular}
    \vspace{2mm}
    \caption{Ranges of physics parameter randomizations.}
\label{table:rand-physics}
\end{table}

\subsubsection{OpenAI Observations}
\label{section-sh-openai-01-tol}

\begin{figure}[h]
\centering
\begin{subfigure}{.5\textwidth}
  \centering
  \includegraphics[width=.9\linewidth]{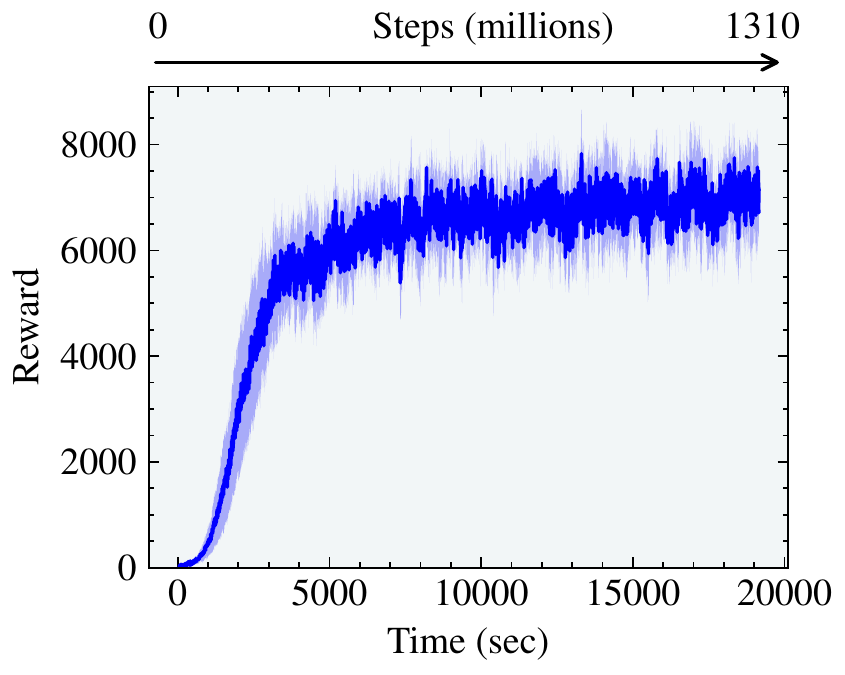}
  \caption{Reward}
  \label{fig:sh-openai-ff-reward}
\end{subfigure}%
\begin{subfigure}{.5\textwidth}
  \centering
  \includegraphics[width=.86\linewidth]{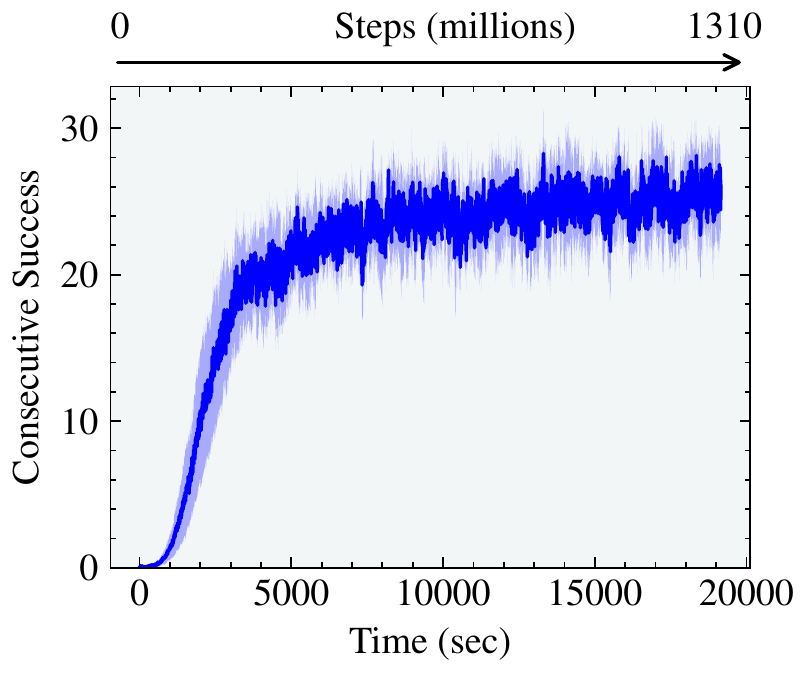}
  \caption{Consecutive Successes}
  \label{fig:sh-openai-ff-success}
\end{subfigure}
\caption{Training curves for ShadowHand environments with \textbf{OpenAI observations} and Feed Forward policy and value functions with a tighter success tolerance of 0.1 rad.}
\label{fig:sh-openai-training-tol-0.1}
\end{figure}

\begin{figure}[h]
\centering
\begin{subfigure}{.5\textwidth}
  \centering
  \includegraphics[width=.9\linewidth]{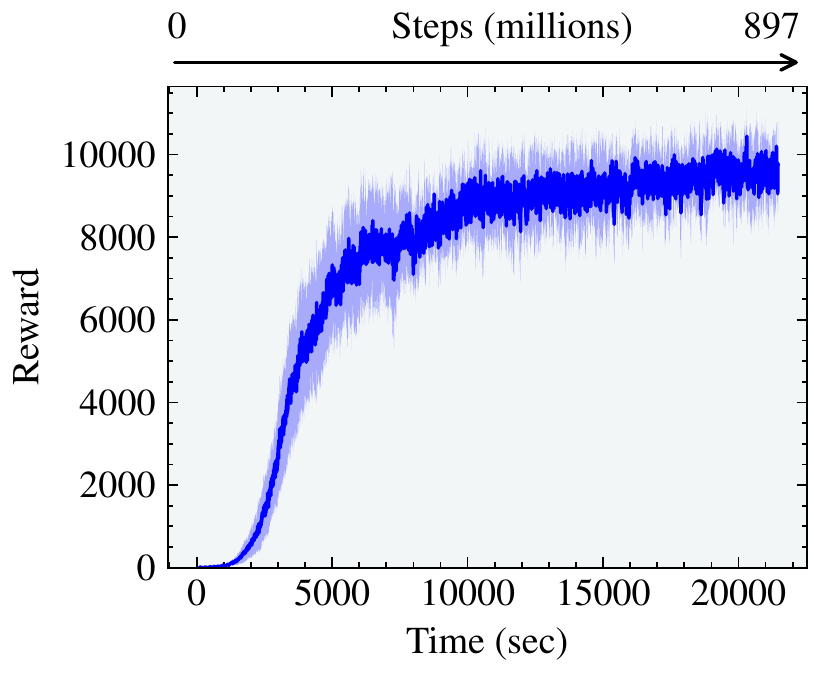}
  \caption{Reward}
  \label{fig:sh-openai-lstm-reward}
\end{subfigure}%
\begin{subfigure}{.5\textwidth}
  \centering
  \includegraphics[width=.86\linewidth]{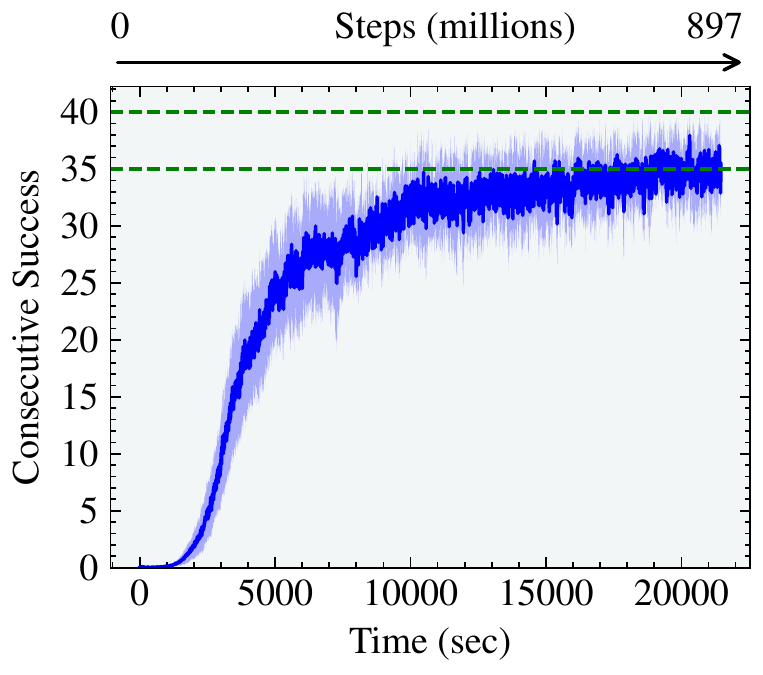}
  \caption{Consecutive Successes}
  \label{fig:sh-openai-lstm-success}
\end{subfigure}
\caption{Training curves for ShadowHand environments with \textbf{OpenAI observations} and an LSTM based policy and value function with a tighter success tolerance of 0.1 rad.}
\label{fig:sh-openai-lstm-training-tol-0.1}
\end{figure}

We conduct experiments with Shadow Hand OpenAI observations with a tighter success tolerance of 0.1 radians and show the reward curves as well as the consecutive successes achieved with this training in Figure \ref{fig:sh-openai-training-tol-0.1} and \ref{fig:sh-openai-lstm-training-tol-0.1}.

\paragraph{Feed Forward Networks}

We achieve 20 consecutive successful cube rotations after training in just under 1 hour. This is similar to the performance\footnote{pp 13, Section 6.5 titled \textbf{Sample Complexity \& Scale}} achieved by OpenAI \textit{et al.} \cite{openai-sh} but with a cluster of 384 16-core CPUs and 8 V100 GPUs with training for 30 hours while we only need a single A100. 

\paragraph{LSTMs} Using sequence networks like LSTMs improve the performance and we find that we are able to achieve 37 consecutive successful cube rotations after training in just under 6 hours. OpenAI \textit{et al.} \cite{openai-sh} achieve similar performance in about 17 hours again on a cluster of 384 16-core CPUs and 8 V100 GPUs. We use a sequence length of 4 to train the LSTM. Various other parameters for this set up are in Table \ref{tab:hyperparameters}. 

We also note that training with a tolerance of 0.1 rad and testing with a tolerance of 0.4 rad, we are able to even go up to 44 consecutive cube rotations.

\end{document}